\documentclass[11pt]{article}

\usepackage{acl}
\usepackage{times}

\usepackage{latexsym}
\usepackage[T1]{fontenc}
\usepackage[utf8]{inputenc}
\usepackage{microtype}
\usepackage{amsmath}
\usepackage{xcolor}
\usepackage{booktabs,multirow,graphicx,colortbl}
\usepackage{amsfonts}
\usepackage{tikz}
\usepackage{amssymb}
\usepackage[most]{tcolorbox}
\usepackage{enumitem}

\usepackage{algorithm}
\usepackage{algpseudocode}

\usepackage{pgfplots,pgfplotstable}
\tikzset{
  font={\fontsize{8pt}{10}\selectfont}}
\usepackage{adjustbox}

\tcbset{
  colframe=gray!30,
  colback=gray!4,
  boxrule=0.4pt,
  arc=2pt,
  outer arc=1pt,
  left=4pt,
  right=4pt,
  top=4pt,
  bottom=4pt,
  fonttitle=\bfseries,
  coltitle=black,
  enhanced jigsaw,
  breakable,
}

\usepackage{arydshln}
\makeatletter
\def\adl@drawiv#1#2#3{%
        \hskip.5\tabcolsep
        \xleaders#3{#2.5\@tempdimb #1{1}#2.5\@tempdimb}%
                #2\z@ plus1fil minus1fil\relax
        \hskip.5\tabcolsep}
\newcommand{\cdashlinelr}[1]{%
  \noalign{\vskip 1.3pt
           \global\let\@dashdrawstore\adl@draw
           \global\let\adl@draw\adl@drawiv}
  \cdashline{#1}[.4pt/2pt]
  \noalign{\global\let\adl@draw\@dashdrawstore
           \vskip 3pt}}
\makeatother

\definecolor{LLMcolor}{HTML}{EE9B51}

\newtcolorbox{promptbox}[2][]{
  title={#2},
  coltitle=black,
  fontupper=\ttfamily\small\sloppy,
  #1
}

\newcommand{\coloredsquare}[1]{%
  \textcolor{#1}{\scalebox{1.10}{$\blacksquare$}}%
}

\DeclareRobustCommand{\diagonalsquare}[3]{%
  \tikz[baseline=-0.05ex, x=1.65ex, y=1.65ex]{%
    \useasboundingbox (0,0) rectangle (1,1);
    \fill[#1] (0,0) -- (1,0) -- (1,1) -- cycle;
    \fill[#2] (0,0) -- (1,1) -- (0.5,1) -- cycle;
    \fill[#3] (0,0) -- (0.5,1) -- (0,1) -- cycle;
  }%
}

\definecolor{colorhex1}{HTML}{f9f4e6} 
\definecolor{colorhex2}{HTML}{f2fbf6} 
\definecolor{colorhex3}{HTML}{e6f1f9}

\definecolor{colorhex1_v2}{HTML}{e9d390} 
\definecolor{colorhex2_v2}{HTML}{c1ecd4} 
\definecolor{colorhex3_v2}{HTML}{94c3e5}

\definecolor{colorhex4_v2}{HTML}{d97f7f} 
\definecolor{colorhex5_v2}{HTML}{e9c2a5} 
\definecolor{colorhex6_v2}{HTML}{b8d7a8}

\definecolor{CustomBlue}{RGB}{57,83,191}

\newcommand{\srclen}{\textsc{SrcLen}}
\newcommand{\sentinel}{\textsc{Sentinel}}
\newcommand{\precomet}{\textsc{PreComet}}
\newcommand{\seqprob}{\textsc{ProbMargin}}
\newcommand{\surprisal}{\textsc{NegLogProb}}
\newcommand{\greedyent}{\textsc{TopkEnt}}

\newcommand{\towertwob}{\textsc{TowerPlus-2B}}

\newcommand{\qwenthreeeone}{\textsc{Qwen3-1.7B}}
\newcommand{\qwenthreeetwo}{\textsc{Qwen3-4B}}
\newcommand{\qwenthreeethree}{\textsc{Qwen3-8B}}
\newcommand{\wmtpp}{\textsc{WMT24++}}

\newcommand{\comet}{\textsc{Comet-22}}
\newcommand{\cometkiwi}{\textsc{CometKiwi-22}}
\newcommand{\charef}{\textsc{ChrF}}
\newcommand{\semdiv}{\textsc{SemDiv}}
\newcommand{\lexdiv}{\textsc{LexDiv}}

\newcommand{\cansurp}{\textsc{AvgNegLogProb}}
\newcommand{\objstab}{\textsc{ObjStab}}

\DeclareMathOperator*{\argmax}{arg\,max}

\setlist{left=0mm,noitemsep} % lets save some space
\title{When is more search useful? An Empirical Study of Search Utility in Quality-Aware Decoding for Machine Translation}

\author{
  Javier Garcia Gilabert$^{1}$ \quad Rico Sennrich$^{1}$ \quad André F. T. Martins$^{2,3,4}$ \\[1mm]
  $^{1}$Department of Computational Linguistics, University of Zurich \\
  $^{2}$Instituto Superior Técnico, Universidade de Lisboa \\
  $^{3}$Instituto de Telecomunicações \quad  $^{4}$ELLIS Unit Lisbon
}

\begin{document}
\maketitle

\begin{abstract}
Quality-Aware Decoding (QAD) improves machine translation through test-time scaling, generating multiple candidate translations and selecting the final output with a quality-aware decoder. Usually the same candidate budget $N$ is spent on every source sentence, assuming that all inputs require the same amount of search. We show that this assumption does not hold and that the quality gain obtained from additional candidates varies across sentences. We study search utility through three quantities: \textit{total scaling gain}, how much quality a sentence can gain from a much larger candidate pool; \textit{marginal scaling gain}, the expected gain from generating one additional batch of candidates; and the \textit{sampling saturation point}, the candidate budget required to realize most of the total scaling gain. We investigate two sources of information for predicting these quantities: signals available before candidate-set expansion, and statistics computed from the candidate pool accumulated during search. We find that sentence-level signals correlate with total scaling gain but are less informative about the sampling saturation point. In contrast, properties of the current candidate pool provide useful signals of marginal scaling gain. Finally, we show that these signals can identify sentences that obtain larger gains when additional search is selectively applied.
\end{abstract}

\section{Introduction}
Increasing test-time computation has proven to be an effective strategy to improve Large Language Model (LLM) performance in downstream tasks \cite{welleck2024decoding,snell2024scaling,muennighoff-etal-2025-s1}. In Machine Translation (MT), Quality-Aware Decoding (QAD) is a form of test-time scaling where we first generate a pool of candidate translations and then select the best one according to a quality-aware decoder. However, the quality of the resulting translation depends on the candidate pool as much as on the quality-aware decoder: a larger pool is more likely to contain a better hypothesis, but it is also more expensive to generate, since the generation cost scales with the number of samples. Despite this, usually the same number of candidates are sampled for every source sentence \cite{rei-etal-2024-tower}, treating search as equally useful for every sentence. 

Prior work has studied how candidate generation and selection affect QAD performance \cite{fernandes-etal-2022-quality, freitag-etal-2023-epsilon}, as well as how to reduce the computational cost of selection \cite{cheng-vlachos-2023-faster,cheng-etal-2025-bayesian}. However, less attention has been paid to how the value of additional search varies across individual source sentences. Some sentences may reach near-optimal quality with only a few candidates, while others continue to benefit as the candidate budget grows.

In this work, we study this sentence-level variation in search utility for two widely used QAD strategies: Minimum Bayes Risk (MBR) decoding \cite{kumar2004minimum,eikema-aziz-2020-map} and quality-based reranking \cite{lee-etal-2021-discriminative}. We distinguish two aspects of search utility. First, \emph{total scaling gain} measures how much quality a sentence can gain by expanding its candidate pool from a single candidate to a large set of candidates. Second, \emph{marginal scaling gain} measures the expected  improvement from generating an additional batch of candidates to extend a given candidate set. These quantities capture complementary signals: a sentence may have substantial overall room for improvement while having already achieved most of that gain in its current budget.

We then investigate which input signals are predictive of these quantities. Finally, we test whether these signals can identify which sentences should be prioritized for additional search. To isolate this question, we use a controlled selective-expansion setting in which all sentences first receive a small candidate pool, and only a subset is expanded.

\paragraph{Contributions.}
We (i) formalize total scaling gain, marginal scaling gain, and sampling saturation point to study how translation quality scales with candidate budget (\S\ref{sec:formalization}); (ii) show empirically that the number of candidates needed varies across sentences (\S\ref{sec:scaling}); (iii) evaluate source-side, model-internal, candidate-pool, and decoder-objective signals as predictors of total and marginal scaling gain (\S\ref{sec:pre-search}, \S\ref{sec:marginal}); and (iv) show that these predictors identify sentences that achieve larger gains from additional search (\S\ref{sec:allocation}).

\section{Background}

\subsection{Quality-Aware Decoding}

Quality-Aware Decoding strategies for MT have proven to generate more human-aligned translations than maximum-a-posteriori (MAP) decoding strategies (e.g. greedy search, beam search) \cite{fernandes-etal-2022-quality}. Common QAD techniques include quality-based reranking \cite{lee-etal-2021-discriminative}, MBR decoding \cite{kumar2004minimum,eikema-aziz-2020-map,muller-sennrich-2021-understanding}, candidate fusion \cite{farinhas-etal-2023-empirical,vernikos-popescu-belis-2024-dont} or self-refinement with reranking \cite{wu2026ladders}. In this work, we focus on reranking and MBR decoding.

Let $p_\theta(y \mid x_i)$ denote the translation model distribution for source sentence $x_i$. We form a set of candidates $C_i^{N_{\max}}=\{y_1,\ldots,y_{N_{\max}}\}$ by sampling $N_{\max}$ translations from this distribution independently.

\paragraph{Minimum Bayes Risk.}
MBR decoding selects the hypothesis that maximizes expected utility under the model distribution according to a quality metric $u(y, y')$, which evaluates the similarity between a candidate $y$ and a reference $y'$:
\begin{equation}
\underset{\tiny \;\;\; y \; \in \; C_i^{N_{\max}} }{\argmax} \; \mathbb{E}_{y' \sim p_\theta(y' \mid x_i)} \left[ u(y, y') \right].
\end{equation}

In MBR, we compute this expectation using each sampled candidate as a pseudo-reference. Because every candidate must be scored pairwise against all $N_{\max}$ pseudo-references, MBR has a quadratic computational complexity of $\mathcal{O}(N_{\max}^2)$.% utility evaluations per source sentence.

\paragraph{Reranking.} In contrast, quality-based reranking evaluates each candidate independently using a reference-free Quality Estimation (QE) metric ${\mathrm{QE}}(x_i, y)$. The selected translation is given by $\arg\max_{y\in C_i^{N_{\max}} } {\mathrm{QE}}(x_i,y)$. Thus, reranking has a linear complexity of $\mathcal{O}(N_{\max})$ evaluations.

\subsection{Reducing the cost of QAD} The computational cost of QAD has motivated several algorithms to make candidate selection more efficient. For MBR decoding, prior work has proposed pruning unlikely hypotheses during utility estimation \cite{cheng-vlachos-2023-faster}, approximating expected utility using clusters of pseudo-references \cite{deguchi-etal-2024-centroid}, aggregating embedding representations of candidates \cite{vamvas-sennrich-2024-linear}, or adaptively allocating utility evaluations across hypotheses \cite{jinnai-ariu-2024-hyperparameter}. These methods reduce the cost of computing the MBR objective while preserving its translation-quality gains. Related work has also proposed efficient methods for quality-based reranking. \citet{cheng-etal-2025-bayesian} frame reranking as a Bayesian optimization problem, iteratively choosing which candidates to evaluate with an expensive quality estimator. These methods primarily reduce the cost of the selection stage while assuming that a fixed candidate pool has already been generated for each source sentence. This leaves candidate generation largely unchanged, even though its cost grows linearly with the number of samples and can become a major component of inference with LLMs. In work parallel to ours, \citet{chen2026quityoureaheadquit} propose \textsc{Quit}, which generates candidates incrementally and stops when recent changes in the decoder score become small, considering both MBR and reranking. \citet{oostermeijer-jones-2026-land} propose \textsc{Land} for reranking, which models source-specific candidate-score distributions and allocates additional samples according to their expected improvement. Thus, both methods design adaptive candidate-generation strategies. Our work instead studies the underlying variation in search utility: how much individual sentences benefit from additional search, how quickly these gains saturate, and which observable signals predict future gains. We consider a broader range of predictors across both MBR and reranking, including source-side and model-internal signals. Unlike \textsc{Quit} and \textsc{Land}, our goal is to characterize and predict when additional search is useful, which we validate through correlational analyses and downstream results with a two-step approach.

\section{Measuring sentence-level search utility}
\label{sec:formalization}

Increasing the number of candidates can improve the selected translation by exposing the quality-aware decoder to better hypotheses \cite{farinhas2024reranking,tan2025investigating}. However, these gains might not be distributed uniformly across source sentences. We formalize this variation below in terms of the gain obtained from additional search and the budget required to achieve it. Consider a QAD method $d$, that selects an output from a sampled candidate set $C_i^{N_{\max}}$. For each source sentence $x_i$, let $C_i^n$ denote a subset of $n$ candidates sampled uniformly without replacement from $C_i^{N_{\max}}$. We define the expected quality obtained with a candidate budget of $n$ as
\begin{equation}
Q_i^d(n)
=
\mathbb{E}_{C_i^n}
\left[
M\bigl(d(x_i,C_i^n), z_i\bigr)
\right].
\end{equation}

where $d(x_i,C_i^n)$ is the output selected by a quality-aware decoder $d$, and $M(\cdot,z_i)$ is a quality metric evaluated against reference $z_i$. We approximate this expectation by repeatedly resampling candidate subsets from the large pool, using $20$ trials. The function $Q_i^d(n)$ thus characterizes how translation quality scales with the candidate budget for sentence $i$ under decoder $d$. Then, we define three metrics that capture different aspects of search utility.

\paragraph{Marginal scaling gain.} 

To model the expansion of an existing candidate pool, we consider nested candidate sets such that $C_i^n \subset C_i^{n+b}$. Given a current budget $n$ and an additional batch of $b$ candidates, we define the marginal scaling gain as

\begin{equation}
\Delta_i^d(n,b)
=
Q_i^d(n+b)
-
Q_i^d(n).
\end{equation}

This quantity measures the expected return from allocating $b$ additional candidates to input $i$ after already spending a budget of $n$ candidates. It corresponds to an online allocation decision.

\paragraph{Total scaling gain.}
We define the total scaling gain as the improvement obtained by expanding the search from an initial budget of one candidate to the largest budget considered

\begin{equation}
    G_i^d
    =
    Q_i^d(N_{\max})
    -
    Q_i^d(1)
    =
    \Delta_i^d(1,N_{\max}-1).
\end{equation}

$G_i^d$ measures the overall amount of quality obtainable from additional search. Inputs with large $G_i^d$ have greater potential to benefit from additional search, whereas a small $G_i^d$ indicates that expanding the candidate set offers little improvement.

\paragraph{Sampling saturation point.} Two inputs with the same total scaling gain can have very different scaling curves. One may achieve most of its available gain with only a few candidates and already be saturated at budget $n$, while another may continue improving as the candidate set grows. For every sentence, we define its sampling saturation point $N_i^\tau$ as the smallest budget required to achieve a fraction $\tau$ of its total scaling gain
\begin{equation}
    N_i^\tau =
    \min \left\{
        n :
        Q_i^d(n) \geq
        Q_i^d(1) + \tau G_i^d
    \right\}.
\end{equation}

\section{Experimental setup}

\subsection{Dataset}

We evaluate on the \wmtpp\ dataset \cite{deutsch-etal-2025-wmt24} using its 960-sentence test set covering three language pairs en$\rightarrow$\{es,ru,de\}.

\subsection{Models}

We experiment with three general-purpose \textsc{Qwen3} models, \qwenthreeeone{}, \qwenthreeetwo{}, and \qwenthreeethree{} \cite{qwen3technicalreport}, as well as \towertwob{}, a multilingual model specialized for machine translation \cite{rei-etal-2026-tower}.
\subsection{Quality-Aware Decoding}

For MBR, candidate utilities are computed with \comet{} \cite{rei-etal-2022-comet}. For reranking, candidates are scored with the reference-free \cometkiwi{} model \cite{rei-etal-2022-cometkiwi}. Model details are provided in Appendix~\ref{app:qad-models}. We use \textsf{\small mbrs} library for MBR and reranking implementation \cite{deguchi-etal-2024-mbrs}.

\subsection{Evaluation metric}

We evaluate translation quality with reference-based \comet{} \cite{rei-etal-2022-comet}, using the references provided in \wmtpp{}. Under MBR, \comet{} is also used as the utility function, albeit scored against sampled pseudo-references rather than the human reference. Our main MBR analysis therefore characterizes \emph{objective-space search utility}: how additional search changes the score under the metric that also guides selection. We study this aligned setting because predicting future improvement in the selection objective is directly relevant to adaptive candidate allocation. However, gains measured in this aligned setting may not fully generalize to other evaluation metrics, as differences between decoding and evaluation objectives can introduce metric interference \cite{pombal-etal-2025-adding}. Thus, in Appendix~\ref{app:decoder-evaluator-alignment} we study how the predictor strength changes between decoder--evaluator combinations. For reranking, the quality-aware decoder is reference-free and differs from the evaluator, giving a decoupled setup.

\subsection{Inference}

We generate 1024 candidates for each source sentence using $\epsilon$-sampling ($\epsilon=0.005$) \cite{hewitt-etal-2022-truncation} following prior work showing its effectiveness with QAD \cite{freitag-etal-2023-epsilon}. For the generation prompt, we follow the official translation prompt from the WMT 2025 General MT Shared Task \cite{kocmi-etal-2025-findings} (see Appendix \ref{appendix:prompt}). We discuss other generation details in Appendix \ref{appendix:generationdetails}.

\begin{table*}[t]
\setlength{\fboxsep}{4.2pt}
\centering
\small
% Requires: \usepackage{booktabs,multirow,graphicx,colortbl,xcolor}
\renewcommand{\arraystretch}{1.15}
\setlength{\tabcolsep}{6pt}

\resizebox{\textwidth}{!}{%
\begin{tabular}{llcccccccccccccccc}
\toprule
 &  & \multicolumn{4}{c}{\textbf{\qwenthreeethree}} & \multicolumn{4}{c}{\textbf{\qwenthreeetwo}} & \multicolumn{4}{c}{\textbf{\qwenthreeeone}} & \multicolumn{4}{c}{\textbf{\towertwob}} \\
\cmidrule(lr){3-6}
\cmidrule(lr){7-10}
\cmidrule(lr){11-14}
\cmidrule(lr){15-18}
 &  & \multicolumn{2}{c}{\textbf{MBR}} & \multicolumn{2}{c}{\textbf{Reranking}} & \multicolumn{2}{c}{\textbf{MBR}} & \multicolumn{2}{c}{\textbf{Reranking}} & \multicolumn{2}{c}{\textbf{MBR}} & \multicolumn{2}{c}{\textbf{Reranking}} & \multicolumn{2}{c}{\textbf{MBR}} & \multicolumn{2}{c}{\textbf{Reranking}} \\
\cmidrule(lr){3-4}
\cmidrule(lr){5-6}
\cmidrule(lr){7-8}
\cmidrule(lr){9-10}
\cmidrule(lr){11-12}
\cmidrule(lr){13-14}
\cmidrule(lr){15-16}
\cmidrule(lr){17-18}
 &  & \shortstack{$\rho_G$} & \shortstack{$\rho_{N^{\text{.95}}}$} & \shortstack{$\rho_G$} & \shortstack{$\rho_{N^{\text{.95}}}$} & \shortstack{$\rho_G$} & \shortstack{$\rho_{N^{\text{.95}}}$} & \shortstack{$\rho_G$} & \shortstack{$\rho_{N^{\text{.95}}}$} & \shortstack{$\rho_G$} & \shortstack{$\rho_{N^{\text{.95}}}$} & \shortstack{$\rho_G$} & \shortstack{$\rho_{N^{\text{.95}}}$} & \shortstack{$\rho_G$} & \shortstack{$\rho_{N^{\text{.95}}}$} & \shortstack{$\rho_G$} & \shortstack{$\rho_{N^{\text{.95}}}$} \\
\midrule

\multirow[c]{5}{*}{\rotatebox[origin=c]{90}{\textbf{Source-only\;\;\;}}} & {\colorbox{colorhex2_v2}{\makebox[1.3cm][l]{\tiny \precomet\textsubscript{diff.}}}} & \cellcolor{green!60} .41 & \cellcolor{green!37} .26 & \cellcolor{green!38} .26 & \cellcolor{green!14} .10 & \cellcolor{green!58} .40 & \cellcolor{green!41} .28 & \cellcolor{green!41} .29 & \cellcolor{green!19} .13 & \cellcolor{green!41} .29 & \cellcolor{green!44} .30 & \cellcolor{green!26} .18 & \cellcolor{green!25} .17 & \cellcolor{green!49} .34 & \cellcolor{green!40} .28 & \cellcolor{green!38} .26 & \cellcolor{green!16} .11 \\

 & {\colorbox{colorhex2_v2}{\makebox[1.3cm][l]{\tiny \precomet\textsubscript{div.}}}} & \cellcolor{green!50} .35 & \cellcolor{green!40} .28 & \cellcolor{green!27} .19 & \cellcolor{green!13} .09 & \cellcolor{green!57} .39 & \cellcolor{green!43} .30 & \cellcolor{green!38} .26 & \cellcolor{green!19} .13 & \cellcolor{green!50} .34 & \cellcolor{green!42} .29 & \cellcolor{green!34} .24 & \cellcolor{green!27} .18 & \cellcolor{green!28} .20 & \cellcolor{green!37} .25 & \cellcolor{green!27} .18 & \cellcolor{green!14} .10 \\

 & {\colorbox{colorhex2_v2}{\makebox[1.3cm][l]{\tiny \sentinel\textsubscript{24}}}} & \cellcolor{green!53} .37 & \cellcolor{green!46} .32 & \cellcolor{green!30} .21 & \cellcolor{green!17} .12 & \cellcolor{green!52} .36 & \cellcolor{green!50} .34 & \cellcolor{green!36} .25 & \cellcolor{green!23} .16 & \cellcolor{green!33} .23 & \cellcolor{green!49} .34 & \cellcolor{green!19} .13 & \cellcolor{green!31} .21 & \cellcolor{green!36} .25 & \cellcolor{green!44} \textbf{.30} & \cellcolor{green!30} .21 & \cellcolor{green!19} .13 \\

 & {\colorbox{colorhex2_v2}{\makebox[1.3cm][l]{\tiny \sentinel\textsubscript{25}}}} & \cellcolor{green!52} .36 & \cellcolor{green!44} .30 & \cellcolor{green!31} .21 & \cellcolor{green!15} .10 & \cellcolor{green!51} .36 & \cellcolor{green!45} .31 & \cellcolor{green!37} .26 & \cellcolor{green!22} .15 & \cellcolor{green!33} .23 & \cellcolor{green!46} .32 & \cellcolor{green!19} .13 & \cellcolor{green!26} .18 & \cellcolor{green!40} .28 & \cellcolor{green!44} .30 & \cellcolor{green!28} .19 & \cellcolor{green!15} .11 \\

 & {\colorbox{colorhex3_v2!80}{\makebox[1.3cm][l]{\tiny \srclen}}} & \cellcolor{green!24} .17 & \cellcolor{green!53} \textbf{.37} & \cellcolor{green!6} .04 & \cellcolor{green!19} \textbf{.13} & \cellcolor{green!32} .22 & \cellcolor{green!53} \textbf{.37} & \cellcolor{green!16} .11 & \cellcolor{green!26} \textbf{.18} & \cellcolor{green!18} .13 & \cellcolor{green!50} \textbf{.34} & \cellcolor{green!9} .06 & \cellcolor{green!33} \textbf{.23} & \cellcolor{red!21} -.14 & \cellcolor{green!38} .26 & \cellcolor{red!6} -.04 & \cellcolor{green!19} \textbf{.13} \\

\addlinespace[0.5mm]
\cdashlinelr{1-18}

\multirow[c]{3}{*}{\rotatebox[origin=c]{90}{\textbf{Internal}}} & {\colorbox{colorhex1_v2}{\makebox[1.3cm][l]{\tiny \seqprob}}} & \cellcolor{green!76} .52 & \cellcolor{green!28} .19 & \cellcolor{green!45} .31 & \cellcolor{green!10} .07 & \cellcolor{green!80} .55 & \cellcolor{green!32} .22 & \cellcolor{green!49} .34 & \cellcolor{green!5} .03 & \cellcolor{green!75} .52 & \cellcolor{green!25} .17 & \cellcolor{green!50} .35 & \cellcolor{green!9} .06 & \cellcolor{green!90} .62 & \cellcolor{green!21} .15 & \cellcolor{green!56} .39 & \cellcolor{green!1} .01 \\

 & {\colorbox{colorhex1_v2}{\makebox[1.3cm][l]{\tiny \surprisal}}} & \cellcolor{green!80} .55 & \cellcolor{green!30} .21 & \cellcolor{green!47} .33 & \cellcolor{green!11} .07 & \cellcolor{green!87} .60 & \cellcolor{green!35} .25 & \cellcolor{green!53} .36 & \cellcolor{green!6} .04 & \cellcolor{green!81} .56 & \cellcolor{green!28} .20 & \cellcolor{green!54} .37 & \cellcolor{green!10} .07 & \cellcolor{green!96} .66 & \cellcolor{green!23} .16 & \cellcolor{green!60} .42 & \cellcolor{green!0} .00 \\

 & {\colorbox{colorhex1_v2}{\makebox[1.3cm][l]{\tiny \greedyent}}} & \cellcolor{green!82} \textbf{.57} & \cellcolor{green!27} .19 & \cellcolor{green!51} \textbf{.35} & \cellcolor{green!11} .08 & \cellcolor{green!91} \textbf{.63} & \cellcolor{green!34} .23 & \cellcolor{green!56} \textbf{.39} & \cellcolor{green!7} .05 & \cellcolor{green!86} \textbf{.59} & \cellcolor{green!30} .20 & \cellcolor{green!58} \textbf{.40} & \cellcolor{green!12} .08 & \cellcolor{green!100} \textbf{.69} & \cellcolor{green!19} .13 & \cellcolor{green!62} \textbf{.43} & \cellcolor{red!0} .00 \\

\bottomrule
\end{tabular}
}

\caption{
Spearman correlation of source- and generation-time predictors with
scaling gain ($\rho_G$) and the sampling saturation point ($\rho_{N^{\text{.95}}}$) under MBR and reranking. Values are averaged across all language pairs.
}
\label{tab:model-predictor-correlations}
\end{table*}

\begin{figure}[t]
    \centering
    \includegraphics[width=\linewidth]{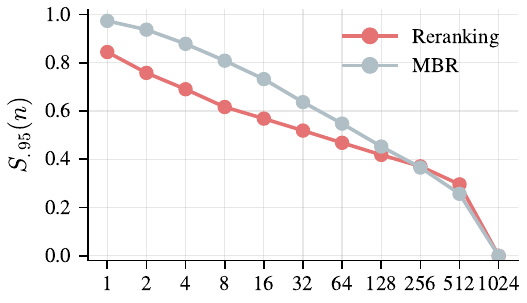}
    \caption{Fraction of sentences that have not yet achieved $95\%$ of their total scaling gain (vertical axis), as a function of candidate budget $n$ (horizontal axis) for \towertwob\ averaged across language pairs.}
    \label{fig:candidate-demand-survival}
\end{figure}

\section{Results and analyses}

We organize our analysis around three questions. First, we show that the sampling saturation point is sentence-dependent. We then study which signals predict total and marginal scaling gain. Finally, we test whether predictions of marginal scaling gain correspond to realized gains.

\subsection{How much search is needed?}
\label{sec:scaling}

We first study how the amount of search required varies across sentences. Using the large candidate pools introduced in Section~\ref{sec:formalization}, we estimate sentence-level quality--budget curves $Q_i^d(n)$ for MBR and reranking over increasing candidate budgets. We summarize the distribution of the sampling saturation point through the survival function

\begin{equation}
    S_\tau(n)
    =
    \frac{
        \#\{i : N_i^\tau > n\}
    }{
        |\mathcal{D}|
    }.
\end{equation}

where $|\mathcal{D}|$ is the total number of sentences in the dataset. $S_\tau(n)$ measures the fraction of sentences that still require more than $n$ candidates to reach a fraction $\tau$ of their total scaling gain.

\paragraph{Not all sentences benefit equally from test-time scaling.}

While the diminishing returns of test-time-scaling are well-known \citep{tan2025investigating}, Figure~\ref{fig:candidate-demand-survival} shows that this is partially because some sentences reach their sampling saturation point much earlier than others.
We report $S_{\text{.95}}(n)$ for \towertwob{} in Figure~\ref{fig:candidate-demand-survival} averaged across language pairs (results by language pair and model can be found in Appendix \ref{app:search-complexity}). Across QAD methods, sentences differ in how much search is required to achieve their available gain. For example, at a budget of 128 candidates about $60\%$ of sentences have already achieved at least $95\%$ of their total scaling gain, whereas many others continue to improve with larger candidate sets. This heterogeneity motivates our next experiments, which asks whether we can predict both \emph{which} sentences have high scaling gain and \emph{how much} search they require.

\subsection{Can we predict total scaling gain?}
\label{sec:pre-search}

We first study predictors that can be computed before expanding the candidate pool, and measure how well they identify sentences with high total
scaling gain $G_i^d$. These signals include source-only properties, predictions from externally trained models, and uncertainty estimates derived from the translation model's own greedy decode.

\paragraph{\coloredsquare{colorhex2_v2} Learned.}
We consider source-only regressors trained to predict properties of a sentence that may be related to its search potential.

\begin{itemize}

    \item \textbf{\sentinel} \cite{proietti-etal-2025-estimating} is an estimator of translation difficulty that predicts the expected quality of a translation using the source sentence only. We include it to test whether sentences that are harder for MT systems also have more room for quality improvement through additional search. We use both released variants, \sentinel\textsubscript{24} and \sentinel\textsubscript{25} and negate their predicted scores so that higher values indicate greater translation difficulty.

    \item \textbf{\precomet} \cite{zouhar-etal-2025-select} is a family of source-based regressors trained to estimate the usefulness of a sentence for MT evaluation. We consider two variants: \precomet\textsubscript{diff.}, which estimates sentence difficulty, and \precomet\textsubscript{div.}, which predicts whether a source sentence is likely to elicit diverse translations across MT systems. We use the difficulty score unchanged and negate the diversity variant's score so that higher values indicate greater expected diversity. 
    
\end{itemize}

\paragraph{\coloredsquare{colorhex3_v2} Heuristic.}
As a simple text-based heuristic, we use the source length (\srclen), measured as the number of whitespace-separated words.

\paragraph{\coloredsquare{colorhex1_v2} Model-internal.}
We compute three signals from the translation model's greedy decoding. All signals are averaged over the generated tokens.
\begin{itemize}

\item \textbf{\greedyent} measures the mean top-$k$ entropy of the next-token distribution, with $k=32$. Token entropy has been previously used as an uncertainty signal in MT \cite{fomicheva-etal-2020-unsupervised}. Following \citet{scalena2026eager}, we restrict the computation to the $k$ most probable tokens to reduce computational overhead when using \textsf{\small vLLM}.

%\andre{how much more expensive would it be to compute the full entropy? this doesn't seem to be a bottleneck. Just computing top-k is probably more expensive than computing the full entropy...}. 

\item \textbf{\seqprob} measures prediction confidence as the probability gap between the two most probable next-token predictions \cite{zhao-etal-2020-active}.

\item \textbf{\surprisal} measures the negative log-probability following \citet{guerreiro-etal-2023-optimal}, where the sequence log-probability of the translation was shown to be predictive of hallucinations.

\end{itemize}

\paragraph{Predictors are more informative about total scaling gain than sampling saturation.}
Table~\ref{tab:model-predictor-correlations} reports Spearman correlations between each predictor and total scaling gain $G_i^d$ and the sampling saturation point $N_i^{.95}$, averaged across language pairs (results by model and language pair are reported in Appendix~\ref{app:marginal-correlations}). Across both MBR and reranking, predictors that correlate with total scaling gain show substantially weaker correlations with the sampling saturation point. This suggests that identifying sentences with greater potential to benefit from additional search is easier than predicting the candidate budget required to realize that gain. Correlations with total scaling gain are also consistently higher for MBR than for reranking across models, indicating that the considered signals capture sentence-level variation in search utility more strongly under MBR decoding.

\paragraph{Predictive strength depends on decoder--evaluator alignment.}

The stronger correlations observed for MBR may partly reflect alignment between the quality-aware decoding objective and the evaluation metric. MBR uses \comet{} as its utility function, which is also used to measure scaling gain, whereas reranking optimizes the reference-free \cometkiwi{} model. As shown in Table~\ref{tab:model-predictor-correlations}, \greedyent{} correlates strongly with total scaling gain for MBR but more moderately for reranking. This suggests that predictors of search potential are more informative when additional search is optimized and evaluated under aligned objectives. We further investigate this effect through a controlled cross-metric analysis with \textsc{xComet} \cite{guerreiro-etal-2024-xcomet} in Appendix~\ref{app:decoder-evaluator-alignment}.

\begin{figure*}[h]
    \centering
    \includegraphics[width=\linewidth]{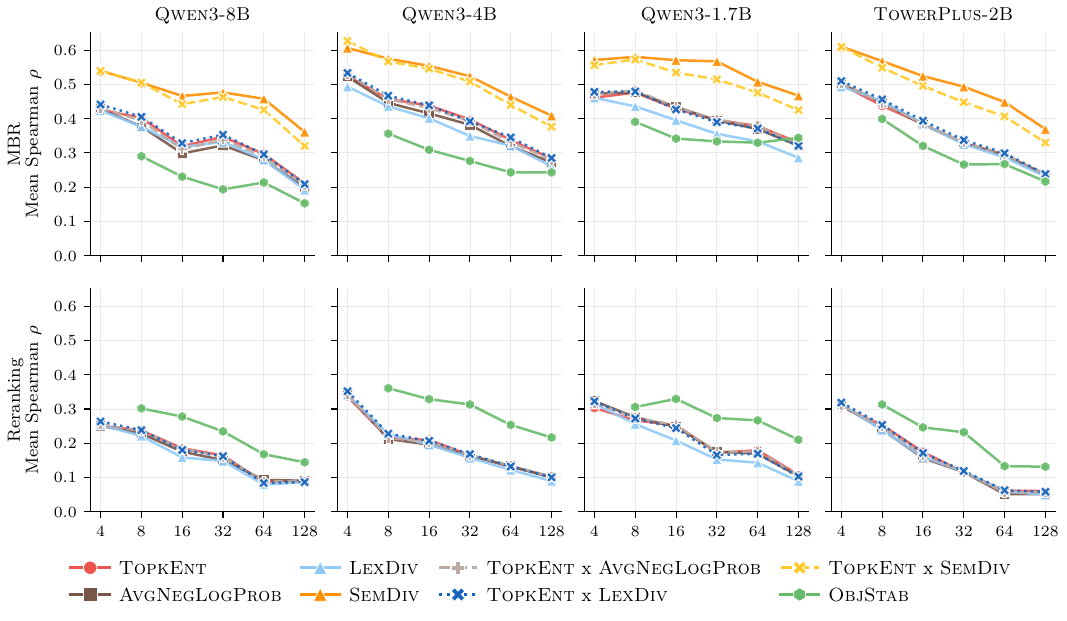}
    \caption{
    Spearman correlation with marginal scaling gain $\Delta_i^d(n,n)$ (vertical axis) as a function of
    candidate budget $n$ (horizontal axis), averaged across language pairs.}
    \label{fig:marginal-corr-vs-budget}
\end{figure*}

\subsection{Can we allocate the full budget at once?}

Predicting $G_i^d$ is not the same as predicting how many candidates a sentence needs, i.e.\ its sampling saturation point $N_i^\tau$ (Section~\ref{sec:scaling}). Allocating the full candidate budget before decoding requires predicting $N_i^\tau$ directly. However, as shown above, predictors are less correlated with $N_i^{.95}$ than with $G_i^d$. This motivates predicting marginal scaling gain, where we consider information about the current candidate pool.

\begin{figure*}[h]
    \centering
    \includegraphics[width=\linewidth]{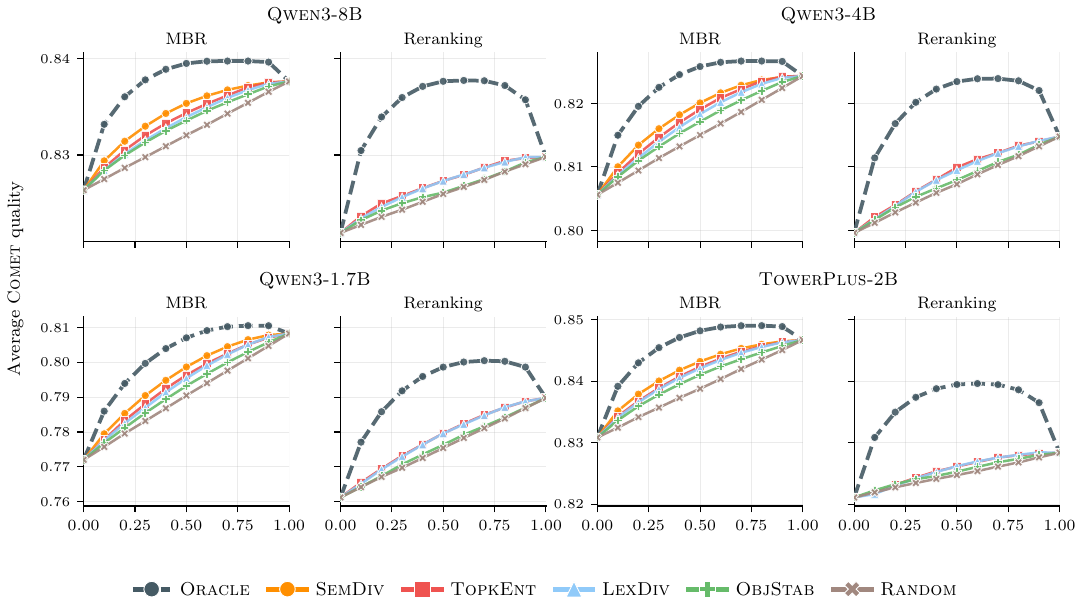}
    \caption{Final dataset-level translation quality (vertical axis) as a function of the fraction $p$ of sentences selected (horizontal axis) from $N_{\mathrm{low}}=8$ to $N_{\mathrm{high}}=256$ candidates. Sentences not selected remain at the initial budget.}
    \label{fig:selective-expansion}
\end{figure*}

\subsection{Predicting marginal scaling gain}
\label{sec:marginal}

Total scaling gain captures whether a sentence can benefit from search, but not how much of that gain remains at the current budget. In this section, we study the expected quality improvement from further expanding the candidate pool. We consider expansions that double the current candidate budget, specifically $n\in\{4,8,16,32,64,128\}$. At each budget $n$, we set the additional batch size to $b=n$ and define marginal scaling gain as follows

\begin{equation}
\Delta_i^d(n,n)
=
Q_i^d(2n)
-
Q_i^d(n).
\end{equation}

Unlike total scaling gain, $\Delta_i^d(n,n)$ depends not only on the sentence's overall scaling gain, but also on the current state of search. Thus, we consider predictors that capture these two signals.

\paragraph{\coloredsquare{colorhex4_v2} Search-potential signal.}

From the predictors studied in Section~\ref{sec:pre-search}, we use {\greedyent}, the strongest predictor of total scaling gain. This quantity remains fixed across candidate budgets.

\paragraph{\coloredsquare{colorhex5_v2} Candidate-pool signals.}
Once $n$ candidates have been generated, we can use properties of the current candidate pool to evaluate search progress.

\begin{itemize}

\item \textbf{\lexdiv} measures lexical diversity within the
current candidate pool using one minus the mean pairwise
\charef\ similarity\footnote{To limit computational overhead, we use all candidate pairs when there are at most 32, and otherwise sample 32.} \cite{popovic-2015-chrf}.

\item \textbf{\semdiv} measures semantic diversity within the
current candidate pool using one minus the mean pairwise
\comet\ similarity \cite{farinhas-etal-2023-empirical}. In MBR decoding, this signal is available at no additional cost from the utility matrix. Thus, we only report it for MBR.

\item \textbf{\cansurp} measures the average \surprisal{} of the candidates in the pool.

\end{itemize}

\paragraph{\coloredsquare{colorhex6_v2} Decoder-objective stabilization.}
Following \citet{chen2026quityoureaheadquit}, we additionally consider {\objstab}, which measures whether the objective used by the quality-aware decoder has stabilized across consecutive candidate budgets. Let $u_i^d(n)$ denote the highest objective score for sentence $i$ at candidate budget $n$. Under our doubling schedule, we define {\objstab} as $\left|u_i^d(n)-u_i^d(n/2)\right|$. For reranking, $u_i^d(n)$ is the highest \textsc{CometKiwi} score among the first $n$ candidates; for MBR, it is the highest expected-utility score. A small value indicates that increasing the candidate pool from $n/2$ to $n$ produced little change in the quality-aware decoder's objective. Unlike candidate-pool signals, {\objstab} directly captures the stabilization of the selection objective. This allows us to evaluate whether objective stabilization indicates that further search is unlikely to improve translation quality. Note that because \objstab\ requires observations at two consecutive candidate budgets, it becomes available only at budget 8.

\paragraph{\diagonalsquare{colorhex4_v2}{colorhex5_v2}{colorhex5_v2} Combining search potential and search state.}
Search potential and candidate pool signals provide complementary information.\;{\greedyent} estimates total scaling gain, whereas candidate-pool and decoder-objective signals describe how search has progressed at the current budget. We therefore evaluate the product between the search-potential signal and candidate-pool signals. We keep \textsc{ObjStab} separate, treating it as an alternative search-state signal based on the decoder's own objective.

\paragraph{Semantic diversity is the strongest predictor of marginal scaling gain for MBR.}
Figure \ref{fig:marginal-corr-vs-budget} reports, at each budget, the Spearman correlation across sentences between $\Delta_i^d(n,n)$ and each predictor (see Appendix \ref{app:marginal-estimation} for implementation details). For MBR, \semdiv{} generally achieves the highest correlations across models and budgets, with \lexdiv{} providing a weaker signal. This suggests that semantic diversity among the generated hypotheses is informative about the gains available from further sampling. Multiplying either diversity measure by \greedyent{} does not consistently improve its correlation. Thus, although model-internal uncertainty is informative about total scaling gain, it provides limited additional information once the state of the current candidate pool is observed. 

\paragraph{Correlations decrease as the candidate budget grows.} At larger budgets, fewer sentences continue to benefit from additional search (see Section~\ref{sec:scaling}), and the remaining marginal gains are smaller and less dispersed across inputs, making their relative ordering more difficult to predict.

\paragraph{Marginal gains are more predictable for MBR.} Correlations are lower for reranking than for MBR. This difference may partly reflect decoder--evaluator alignment (see Section~\ref{sec:pre-search}). Candidate-pool signals may also be more informative for MBR because its decoding objective explicitly models interactions within the support set, unlike reranking, which scores candidates independently.

\paragraph{\objstab{} is more informative for reranking.} \objstab{} performance differs between decoding methods. It is among the strongest predictors for reranking, but underperforms for MBR. In reranking, changes in the selected hypothesis are tied to improvements in the highest objective score, making its recent stabilization a useful, albeit modest, signal of marginal gain. For MBR, stabilization of the best expected utility captures less information than the semantic structure of the support set, which is better reflected by diversity predictors.

\subsection{Can we identify useful expansions?}
\label{sec:allocation}

The previous section showed that sentence-level signals can predict marginal scaling gain at a given candidate budget. We now ask a different question: \emph{do sentences ranked as having greater future search utility actually obtain larger gains when their candidate pools are expanded?} We consider a controlled one-shot allocation setting that isolates this question. Every sentence first receives an initial candidate budget $N_{\mathrm{low}}$. We then rank sentences according to a predictor of scaling gain and expand the top-$p$ fraction to $N_{\mathrm{high}}>N_{\mathrm{low}}$, while the remaining sentences stay at the initial budget. This allows us to assess whether the signals studied above remain useful when prioritizing sentences for further search. We evaluate signals representing the different sources of information studied in Section~\ref{sec:marginal}. Specifically, we consider \greedyent{} as a fixed estimate of search potential, \lexdiv{} and \semdiv{} as measures of the state of the current candidate pool, and \objstab{} as a measure of decoder-objective stabilization. We set \(N_{\mathrm{low}}=8\), so that \objstab{} can be computed from the decoder objectives at budgets 4 and 8. We compare predictor-based selection against \textsc{Random}, which selects a fraction \(p\) of sentences uniformly at random, and \textsc{Oracle}, an upper bound that ranks sentences according to their true future gain.

\paragraph{Semantic diversity identifies useful expansions.} Figure \ref{fig:selective-expansion} reports final translation quality as a function of the fraction $p$ of sentences expanded for $N_{\mathrm{high}}=256$ (additional budgets are reported in Appendix \ref{app:selective-expansion}). For MBR, \semdiv{} produces the curve closest to the oracle, followed by \greedyent{} and \lexdiv{}, and gives the largest improvements over random selection. This is consistent with the stronger correlations with marginal scaling gain observed in Section~\ref{sec:marginal} for MBR. The oracle curves show greater potential for improvement, suggesting that having better estimates of future search utility could further improve selective expansion.

\paragraph{Objective stabilization is less informative for selective expansion.} Although \objstab{} correlates with gains at consecutive budgets, it performs worse in selective expansion. This contrasts with \textsc{Quit} \cite{chen2026quityoureaheadquit}, where objective stabilization is used for local stopping, suggesting that local objective convergence alone is insufficient for estimating longer-horizon gains.

\begin{figure}[h]
    \centering
    \includegraphics[width=\linewidth]{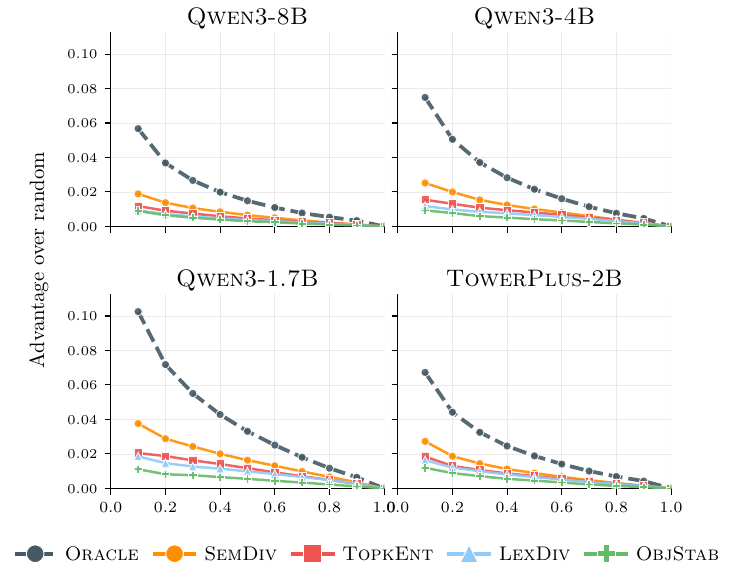}
    \caption{ Advantage over random selection in average marginal scaling gain (vertical axis) as a function of the fraction \(p\) of selected sentences (horizontal axis).}
    \label{fig:selected-gain}
\end{figure}

\paragraph{Semantic diversity-based rankings better identify future search utility.} To understand the source of these improvements, we focus on MBR, where predictor-based selection shows the clearest gains over random selection. Figure \ref{fig:selected-gain} reports the advantage over random in the average future gain of the top-\(p\) fraction selected for expansion in MBR. \semdiv{} achieves the largest advantage showing that diversity-based rankings select sentences for which the larger candidate pool is more useful. The advantage is greatest at small values of \(p\) and decreases as more sentences are included. Appendix~\ref{app:selected-gain} reports absolute gains by model and language pair.

\section{Conclusion and Future Work}
\label{sec:conclusion}

We studied when additional search is useful in QAD. We showed that the amount of search required varies across sentences and formalized this variation through total scaling gain, marginal scaling gain, and the sampling saturation point. Signals that correlate with total scaling gain are considerably less informative about sampling saturation, suggesting that identifying sentences with greater potential to benefit from additional search is easier than predicting how much search they require. During candidate generation, candidate-pool signals are more informative about marginal scaling gain. For MBR, semantic diversity is the most consistent predictor across models and candidate budgets. Finally, selective-expansion experiments show that studied signals can identify sentences that obtain larger gains from additional search than random selection, especially for MBR. Overall, our results suggest that sentence-level search utility is predictable, but that its predictability differs substantially across decoding methods. Future work could incorporate such signals into bandit-based dynamic allocation algorithms \citep{wang2025dynscaling}.

\section*{Limitations}

First, our definition of scaling gain is metric-dependent. In our main MBR experiments, \comet{} is used both as the decoding utility and the evaluation metric. Accordingly, these experiments characterize \emph{objective-space search utility}: how predictably additional search improves the metric that guides candidate selection. Our cross-metric analysis in Appendix~\ref{app:decoder-evaluator-alignment} shows that predictor strength varies across decoder--evaluator combinations. More generally, optimizing and evaluating with related metrics can introduce metric interference, a known issue in metric-based decoding \cite{pombal-etal-2025-adding}. We account for this by explicitly studying decoder--evaluator alignment, while noting that our main conclusions concern the relative predictability of search utility under the specified objective. Thus, the absolute correlations should be interpreted with respect to the metric used to define search utility.

Second, our experiments cover three translation directions, four models, one sampling strategy, and two QAD methods. The observed relationships may differ for other language pairs, model families, domains, sampling distributions, or metrics.

Lastly, our selective-expansion experiments are designed as a controlled diagnostic of whether the proposed signals rank sentences according to their future search utility. They are not intended to establish that selective expansion is more compute-efficient than uniform allocation, nor to identify an optimal allocation policy. In particular, we consider a fixed two-level expansion from \(N_{\mathrm{low}}\) to \(N_{\mathrm{high}}\), rather than comparing alternative ways of distributing the same inference budget across sentences. Developing and evaluating end-to-end dynamic allocation strategies under matched computational budgets is left for future work. Despite this, we believe our findings hold significance for the MT community.

\section*{Ethical considerations}

Quality-aware decoding selects among candidate translations using automatic metrics, and may therefore inherit the biases encoded in those metrics. Prior work shows that quality estimation metrics can exhibit gender bias, and that this bias can propagate to quality-aware decoding by systematically favoring certain translation forms over others \citep{zaranis-etal-2025-watching}. Both the underlying language models and the metrics used for decoding and evaluation may encode social biases that shape which candidates are generated or preferred. We do not study how search and quality-aware decoding interact with these biases, and we leave this important question to future work.

\section*{Generative AI Use Disclosure}

We used generative AI tools to help draft and polish parts of the text, write code, and prepare figures. The authors developed and verified all scientific content, analyzes and conclusions.

\section*{Acknowledgments}

This work was funded by the Swiss National Science Foundation (project TINE, 20CH-1\_238345). 
AM was supported by the project DECOLLAGE (ERC-2022-CoG
101088763) and by FCT/MECI through national funds and when applicable co-funded EU funds
under UID/50008: Instituto de Telecomunicações. 

\bibliography{custom}

\appendix

\section{Prompt}
\label{appendix:prompt}

\begin{promptbox}[colback=LLMcolor!20!white, colframe=LLMcolor!80]{LLMs Prompt}
You are a professional \textbf{\{src\_lang\}}-to-\textbf{\{tgt\_lang\}} translator. Your goal is to accurately convey
the meaning and nuances of the original \textbf{\{src\_lang\}} text while adhering to \textbf{\{tgt\_lang\}} grammar,
vocabulary, and cultural sensitivities. Preserve the line breaks. Use precise terminology
and a tone appropriate for academic or instructional materials. Produce only the \textbf{\{tgt\_lang\}}
translation, without any additional explanations or commentary. Please translate the
provided \textbf{\{src\_lang\}} text into \textbf{\{tgt\_lang\}}:
\end{promptbox}

\section{Quality-Aware decoding models}
\label{app:qad-models}

For MBR decoding, we use the \texttt{Unbabel/wmt22-comet-da} model for candidate utility computation. For quality-based reranking, we use the reference-free \texttt{Unbabel/wmt22-cometkiwi-da} model.

\begin{itemize}
    \item \comet{}: \url{https://huggingface.co/Unbabel/wmt22-comet-da}
    \item \cometkiwi{}: \url{https://huggingface.co/Unbabel/wmt22-cometkiwi-da}
\end{itemize}

\section{Generation details}
\label{appendix:generationdetails}

We use \textsf{\small vLLM v0.10.2} \cite{kwon2023efficient} for inference, with model weights loaded in \texttt{bfloat16} precision. Generation is performed on a server equipped with eight NVIDIA GeForce RTX 4090 GPUs, each with 24\,GB of memory. The server uses NVIDIA driver version \texttt{570.124.06}, which supports CUDA~12.8.

\section{Sampling saturation point across models}
\label{app:search-complexity}

\begin{figure*}[!h]
    \centering

    \towertwob
    
    \includegraphics[width=\linewidth]{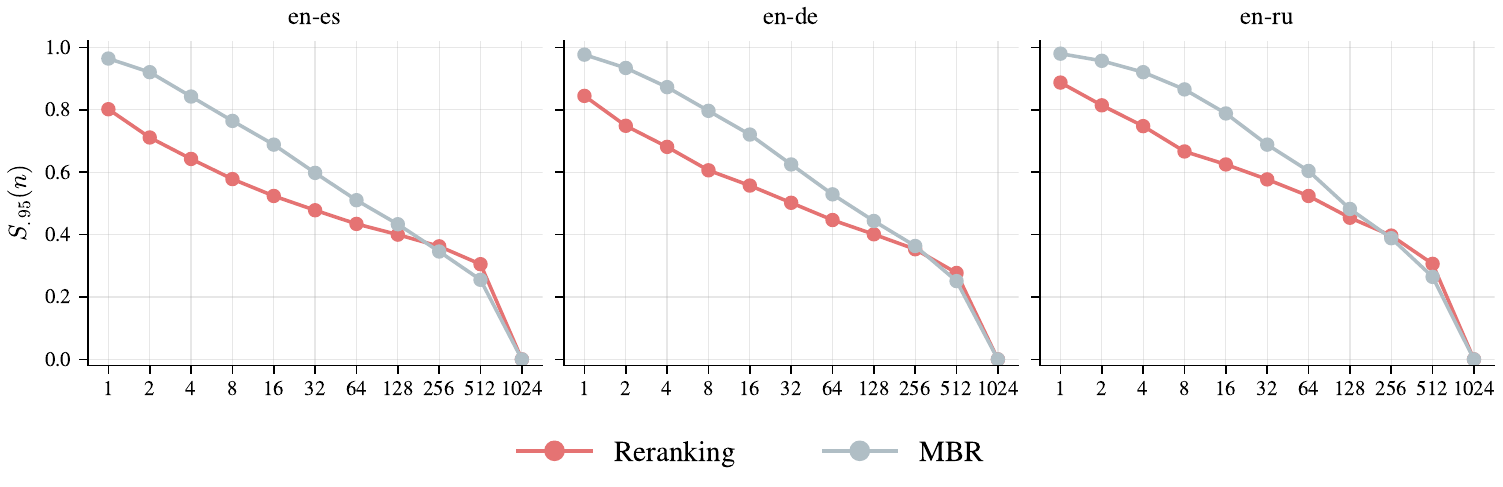}

    \qwenthreeeone
    
    \includegraphics[width=\linewidth]{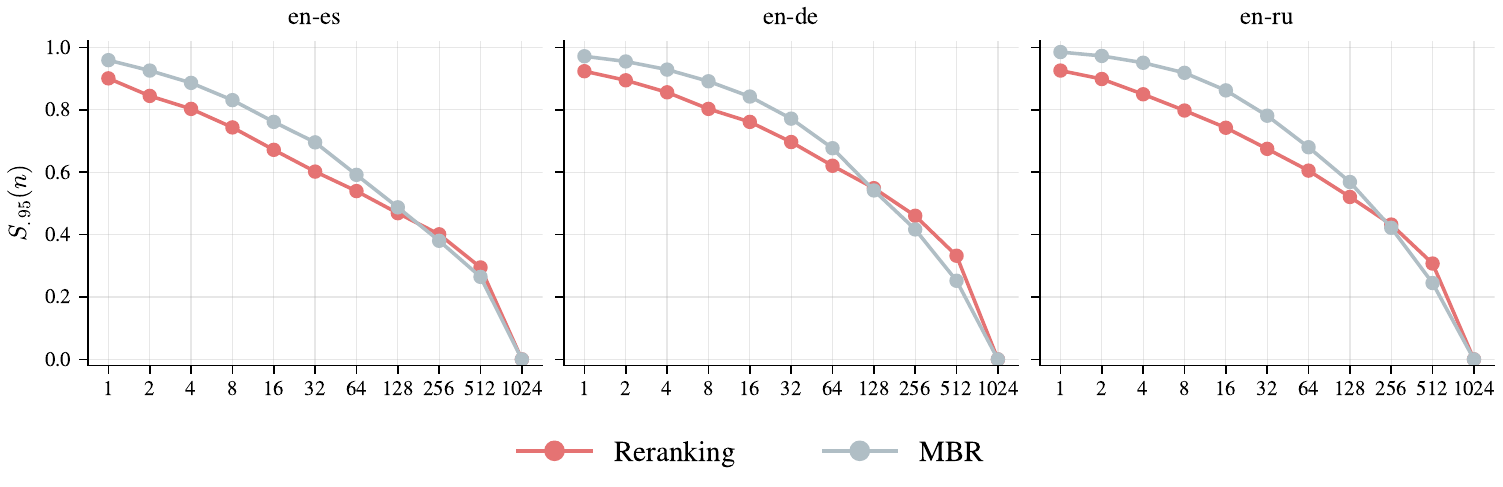}

    \vspace{0.8em}

    \qwenthreeetwo
    
    \includegraphics[width=\linewidth]{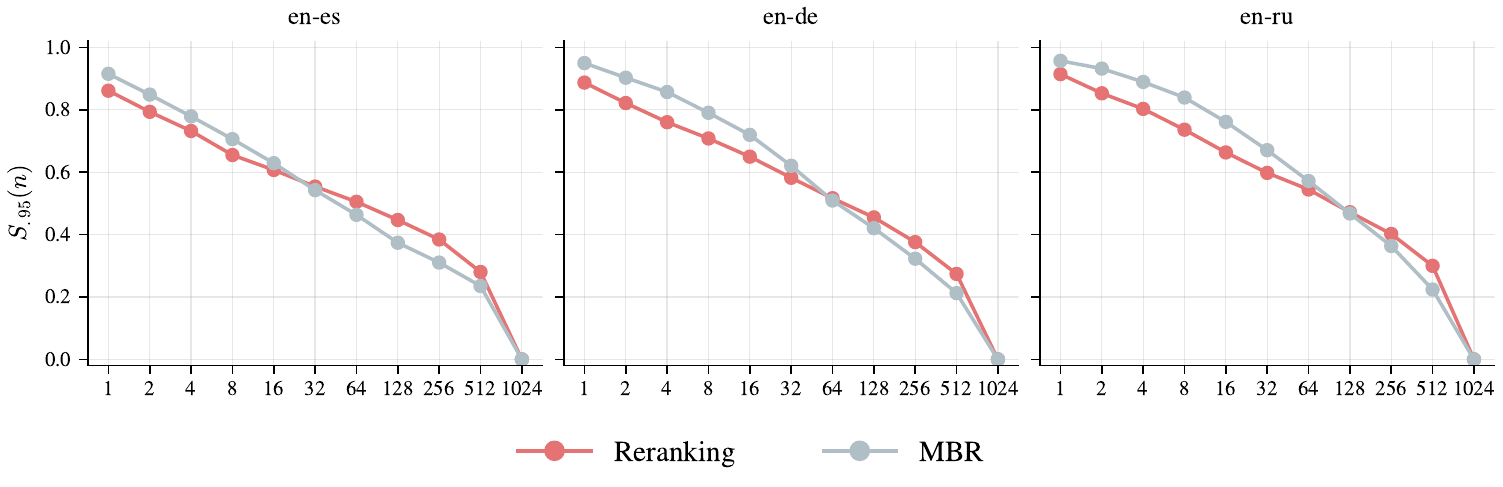}

    \vspace{0.8em}

    \qwenthreeethree
    
    \includegraphics[width=\linewidth]{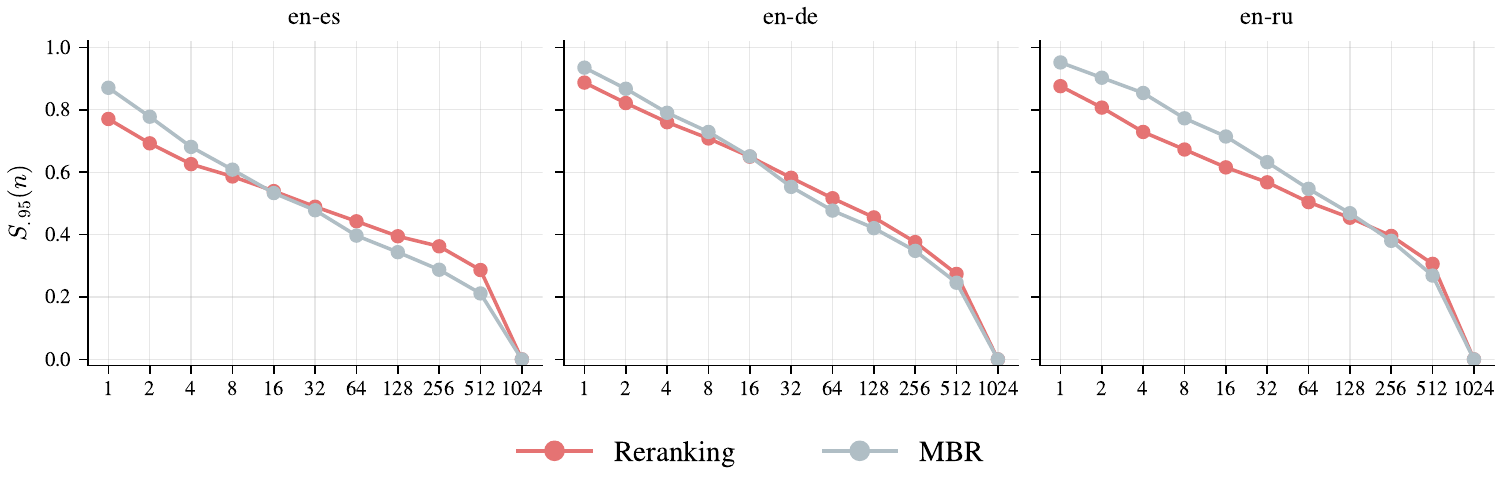}

    \caption{
    Fraction of sentences that have not yet achieved $95\%$ of their total scaling gain (vertical axis), as a function of candidate budget $n$ (horizontal axis).
    }

    \label{fig:candidate-demand-survival-qwen3}
\end{figure*}

We show in Figure \ref{fig:candidate-demand-survival-qwen3} the survival function defined in Section \ref{sec:scaling} for $\tau = 0.95$ for \qwenthreeeone, \qwenthreeetwo{} and \qwenthreeethree{} models.

\section{Search-utility correlations by language pair}
\label{app:marginal-correlations}

Table~\ref{tab:predictor-correlations-lang-pairs} reports the correlations with total scaling gain and the sampling saturation point for each language pair and each model.

\begin{table*}[!h]
\centering
\scriptsize
\renewcommand{\arraystretch}{1.0}
\setlength{\tabcolsep}{6.4pt}

\begin{tabular}{llcccccccccccc}
\toprule

 &  & \multicolumn{4}{c}{\textbf{en-es}}
     & \multicolumn{4}{c}{\textbf{en-de}}
     & \multicolumn{4}{c}{\textbf{en-ru}} \\

\cmidrule(lr){3-6}
\cmidrule(lr){7-10}
\cmidrule(lr){11-14}

 &  & \multicolumn{2}{c}{\textbf{MBR}}
     & \multicolumn{2}{c}{\textbf{Reranking}}
     & \multicolumn{2}{c}{\textbf{MBR}}
     & \multicolumn{2}{c}{\textbf{Reranking}}
     & \multicolumn{2}{c}{\textbf{MBR}}
     & \multicolumn{2}{c}{\textbf{Reranking}} \\

\cmidrule(lr){3-4}
\cmidrule(lr){5-6}
\cmidrule(lr){7-8}
\cmidrule(lr){9-10}
\cmidrule(lr){11-12}
\cmidrule(lr){13-14}

 & &
$\rho_G$ & $\rho_{N^{\text{.95}}}$
& $\rho_G$ & $\rho_{N^{\text{.95}}}$
& $\rho_G$ & $\rho_{N^{\text{.95}}}$
& $\rho_G$ & $\rho_{N^{\text{.95}}}$
& $\rho_G$ & $\rho_{N^{\text{.95}}}$
& $\rho_G$ & $\rho_{N^{\text{.95}}}$ \\

\midrule

\multicolumn{14}{c}{\textbf{\qwenthreeethree}} \\
\midrule

\multirow[c]{5}{*}{\rotatebox[origin=c]{90}{\textbf{Source-only\;\;\;}}} & {\colorbox{colorhex2_v2}{\makebox[1.3cm][l]{\tiny \precomet\textsubscript{diff.}}}}  & \cellcolor{green!69} 0.43 & \cellcolor{green!41} 0.26 & \cellcolor{green!47} 0.29 & \cellcolor{green!21} 0.13 & \cellcolor{green!66} 0.41 & \cellcolor{green!38} 0.24 & \cellcolor{green!44} 0.27 & \cellcolor{green!15} 0.10 & \cellcolor{green!64} 0.40 & \cellcolor{green!44} 0.28 & \cellcolor{green!35} 0.22 & \cellcolor{green!11} 0.07 \\

 & {\colorbox{colorhex2_v2}{\makebox[1.3cm][l]{\tiny \precomet\textsubscript{div.}}}} & \cellcolor{green!60} 0.37 & \cellcolor{green!49} 0.31 & \cellcolor{green!30} 0.19 & \cellcolor{green!14} 0.09 & \cellcolor{green!59} 0.37 & \cellcolor{green!44} 0.28 & \cellcolor{green!39} 0.25 & \cellcolor{green!11} 0.07 & \cellcolor{green!47} 0.30 & \cellcolor{green!39} 0.25 & \cellcolor{green!19} 0.12 & \cellcolor{green!18} 0.11 \\

 & {\colorbox{colorhex2_v2}{\makebox[1.3cm][l]{\tiny \sentinel\textsubscript{24}}}} & \cellcolor{green!68} 0.42 & \cellcolor{green!58} 0.36 & \cellcolor{green!42} 0.26 & \cellcolor{green!22} 0.14 & \cellcolor{green!54} 0.34 & \cellcolor{green!48} 0.30 & \cellcolor{green!34} 0.21 & \cellcolor{green!22} 0.14 & \cellcolor{green!54} 0.34 & \cellcolor{green!48} 0.30 & \cellcolor{green!24} 0.15 & \cellcolor{green!12} 0.08 \\

 & {\colorbox{colorhex2_v2}{\makebox[1.3cm][l]{\tiny \sentinel\textsubscript{25}}}} & \cellcolor{green!67} 0.42 & \cellcolor{green!59} 0.37 & \cellcolor{green!42} 0.26 & \cellcolor{green!18} 0.11 & \cellcolor{green!54} 0.34 & \cellcolor{green!42} 0.26 & \cellcolor{green!36} 0.22 & \cellcolor{green!21} 0.13 & \cellcolor{green!53} 0.33 & \cellcolor{green!44} 0.28 & \cellcolor{green!25} 0.16 & \cellcolor{green!10} 0.06 \\

 & {\colorbox{colorhex3_v2!80}{\makebox[1.3cm][l]{\tiny \srclen}}} & \cellcolor{green!43} 0.27 & \cellcolor{green!70} 0.43 & \cellcolor{green!12} 0.08 & \cellcolor{green!21} 0.13 & \cellcolor{green!26} 0.16 & \cellcolor{green!56} 0.35 & \cellcolor{green!16} 0.10 & \cellcolor{green!27} 0.17 & \cellcolor{green!12} 0.08 & \cellcolor{green!52} 0.32 & \cellcolor{red!9} -0.05 & \cellcolor{green!17} 0.11 \\

\addlinespace[0.6mm]
\cdashlinelr{1-14}

\multirow[c]{3}{*}{\rotatebox[origin=c]{90}{\;\;\;\; \textbf{Internal}}} & {\colorbox{colorhex1_v2}{\makebox[1.3cm][l]{\tiny \seqprob}}} & \cellcolor{green!75} 0.47 & \cellcolor{green!33} 0.21 & \cellcolor{green!44} 0.27 & \cellcolor{green!14} 0.09 & \cellcolor{green!91} 0.57 & \cellcolor{green!31} 0.19 & \cellcolor{green!59} 0.37 & \cellcolor{green!20} 0.13 & \cellcolor{green!85} 0.53 & \cellcolor{green!29} 0.18 & \cellcolor{green!48} 0.30 & \cellcolor{green!1} 0.01 \\

 & {\colorbox{colorhex1_v2}{\makebox[1.3cm][l]{\tiny \surprisal}}} & \cellcolor{green!79} 0.49 & \cellcolor{green!34} 0.21 & \cellcolor{green!44} 0.28 & \cellcolor{green!13} 0.08 & \cellcolor{green!96} 0.60 & \cellcolor{green!36} 0.22 & \cellcolor{green!63} 0.39 & \cellcolor{green!21} 0.13 & \cellcolor{green!91} 0.57 & \cellcolor{green!30} 0.19 & \cellcolor{green!50} 0.31 & \cellcolor{green!1} 0.01 \\

 & {\colorbox{colorhex1_v2}{\makebox[1.3cm][l]{\tiny \greedyent}}} & \cellcolor{green!80} 0.50 & \cellcolor{green!28} 0.18 & \cellcolor{green!48} 0.30 & \cellcolor{green!14} 0.09 & \cellcolor{green!100} 0.62 & \cellcolor{green!32} 0.20 & \cellcolor{green!66} 0.41 & \cellcolor{green!20} 0.12 & \cellcolor{green!93} 0.58 & \cellcolor{green!29} 0.18 & \cellcolor{green!54} 0.34 & \cellcolor{green!2} 0.01 \\

\midrule
\multicolumn{14}{c}{\textbf{\qwenthreeetwo}} \\
\midrule

\multirow[c]{5}{*}{\rotatebox[origin=c]{90}{\textbf{Source-only\;\;\;}}} & {\colorbox{colorhex2_v2}{\makebox[1.3cm][l]{\tiny \precomet\textsubscript{diff.}}}} & \cellcolor{green!61} 0.40 & \cellcolor{green!37} 0.24 & \cellcolor{green!42} 0.28 & \cellcolor{green!11} 0.07 & \cellcolor{green!60} 0.40 & \cellcolor{green!52} 0.34 & \cellcolor{green!43} 0.29 & \cellcolor{green!21} 0.14 & \cellcolor{green!61} 0.40 & \cellcolor{green!40} 0.26 & \cellcolor{green!44} 0.29 & \cellcolor{green!29} 0.19 \\

 & {\colorbox{colorhex2_v2}{\makebox[1.3cm][l]{\tiny \precomet\textsubscript{div.}}}} & \cellcolor{green!58} 0.38 & \cellcolor{green!40} 0.27 & \cellcolor{green!34} 0.22 & \cellcolor{green!12} 0.08 & \cellcolor{green!59} 0.39 & \cellcolor{green!53} 0.35 & \cellcolor{green!45} 0.30 & \cellcolor{green!21} 0.14 & \cellcolor{green!62} 0.41 & \cellcolor{green!41} 0.27 & \cellcolor{green!42} 0.28 & \cellcolor{green!26} 0.17 \\

 & {\colorbox{colorhex2_v2}{\makebox[1.3cm][l]{\tiny \sentinel\textsubscript{24}}}} & \cellcolor{green!58} 0.38 & \cellcolor{green!50} 0.33 & \cellcolor{green!40} 0.26 & \cellcolor{green!19} 0.12 & \cellcolor{green!53} 0.35 & \cellcolor{green!58} 0.38 & \cellcolor{green!38} 0.25 & \cellcolor{green!26} 0.17 & \cellcolor{green!54} 0.36 & \cellcolor{green!49} 0.32 & \cellcolor{green!34} 0.23 & \cellcolor{green!28} 0.19 \\

 & {\colorbox{colorhex2_v2}{\makebox[1.3cm][l]{\tiny \sentinel\textsubscript{25}}}} & \cellcolor{green!59} 0.39 & \cellcolor{green!45} 0.30 & \cellcolor{green!44} 0.29 & \cellcolor{green!20} 0.13 & \cellcolor{green!52} 0.34 & \cellcolor{green!55} 0.36 & \cellcolor{green!39} 0.26 & \cellcolor{green!27} 0.18 & \cellcolor{green!51} 0.34 & \cellcolor{green!41} 0.27 & \cellcolor{green!34} 0.22 & \cellcolor{green!23} 0.15 \\

 &  {\colorbox{colorhex3_v2!80}{\makebox[1.3cm][l]{\tiny \srclen}}}  & \cellcolor{green!35} 0.23 & \cellcolor{green!53} 0.35 & \cellcolor{green!11} 0.07 & \cellcolor{green!24} 0.16 & \cellcolor{green!34} 0.22 & \cellcolor{green!64} 0.42 & \cellcolor{green!21} 0.14 & \cellcolor{green!26} 0.17 & \cellcolor{green!33} 0.22 & \cellcolor{green!50} 0.33 & \cellcolor{green!18} 0.12 & \cellcolor{green!32} 0.21 \\

\addlinespace[0.6mm]
\cdashlinelr{1-14}

\multirow[c]{3}{*}{\rotatebox[origin=c]{90}{\;\;\;\; \textbf{Internal}}} & {\colorbox{colorhex1_v2}{\makebox[1.3cm][l]{\tiny \seqprob}}} & \cellcolor{green!84} 0.55 & \cellcolor{green!35} 0.23 & \cellcolor{green!47} 0.31 & \cellcolor{red!3} -0.02 & \cellcolor{green!89} 0.59 & \cellcolor{green!41} 0.27 & \cellcolor{green!53} 0.35 & \cellcolor{green!1} 0.01 & \cellcolor{green!80} 0.53 & \cellcolor{green!25} 0.17 & \cellcolor{green!53} 0.35 & \cellcolor{green!17} 0.11 \\

 & {\colorbox{colorhex1_v2}{\makebox[1.3cm][l]{\tiny \surprisal}}} & \cellcolor{green!89} 0.58 & \cellcolor{green!38} 0.25 & \cellcolor{green!48} 0.32 & \cellcolor{red!2} -0.02 & \cellcolor{green!96} 0.63 & \cellcolor{green!44} 0.29 & \cellcolor{green!57} 0.38 & \cellcolor{green!3} 0.02 & \cellcolor{green!89} 0.59 & \cellcolor{green!30} 0.20 & \cellcolor{green!61} 0.40 & \cellcolor{green!19} 0.13 \\

 & {\colorbox{colorhex1_v2}{\makebox[1.3cm][l]{\tiny \greedyent}}} & \cellcolor{green!93} 0.61 & \cellcolor{green!36} 0.23 & \cellcolor{green!51} 0.34 & \cellcolor{red!3} -0.02 & \cellcolor{green!100} 0.66 & \cellcolor{green!43} 0.28 & \cellcolor{green!62} 0.41 & \cellcolor{green!5} 0.03 & \cellcolor{green!93} 0.61 & \cellcolor{green!28} 0.18 & \cellcolor{green!64} 0.42 & \cellcolor{green!20} 0.13 \\

\midrule
\multicolumn{14}{c}{\textbf{\qwenthreeeone}} \\
\midrule

\multirow[c]{5}{*}{\rotatebox[origin=c]{90}{\textbf{Source-only\;\;\;}}} & {\colorbox{colorhex2_v2}{\makebox[1.3cm][l]{\tiny \precomet\textsubscript{diff.}}}} & \cellcolor{green!52} 0.33 & \cellcolor{green!44} 0.28 & \cellcolor{green!38} 0.24 & \cellcolor{green!27} 0.17 & \cellcolor{green!42} 0.27 & \cellcolor{green!49} 0.31 & \cellcolor{green!28} 0.18 & \cellcolor{green!24} 0.15 & \cellcolor{green!40} 0.25 & \cellcolor{green!49} 0.31 & \cellcolor{green!20} 0.13 & \cellcolor{green!30} 0.19 \\

 & {\colorbox{colorhex2_v2}{\makebox[1.3cm][l]{\tiny \precomet\textsubscript{div.}}}} & \cellcolor{green!52} 0.33 & \cellcolor{green!37} 0.24 & \cellcolor{green!40} 0.25 & \cellcolor{green!27} 0.17 & \cellcolor{green!55} 0.35 & \cellcolor{green!47} 0.30 & \cellcolor{green!35} 0.22 & \cellcolor{green!23} 0.14 & \cellcolor{green!55} 0.35 & \cellcolor{green!51} 0.33 & \cellcolor{green!36} 0.23 & \cellcolor{green!37} 0.24 \\

 & {\colorbox{colorhex2_v2}{\makebox[1.3cm][l]{\tiny \sentinel\textsubscript{24}}}} & \cellcolor{green!41} 0.26 & \cellcolor{green!50} 0.32 & \cellcolor{green!28} 0.18 & \cellcolor{green!29} 0.18 & \cellcolor{green!34} 0.22 & \cellcolor{green!54} 0.34 & \cellcolor{green!22} 0.14 & \cellcolor{green!35} 0.22 & \cellcolor{green!33} 0.21 & \cellcolor{green!57} 0.36 & \cellcolor{green!11} 0.07 & \cellcolor{green!37} 0.23 \\

 & {\colorbox{colorhex2_v2}{\makebox[1.3cm][l]{\tiny \sentinel\textsubscript{25}}}} & \cellcolor{green!40} 0.25 & \cellcolor{green!48} 0.30 & \cellcolor{green!28} 0.18 & \cellcolor{green!24} 0.16 & \cellcolor{green!36} 0.23 & \cellcolor{green!51} 0.33 & \cellcolor{green!26} 0.16 & \cellcolor{green!33} 0.21 & \cellcolor{green!32} 0.20 & \cellcolor{green!50} 0.32 & \cellcolor{green!8} 0.05 & \cellcolor{green!28} 0.18 \\

 & {\colorbox{colorhex3_v2!80}{\makebox[1.3cm][l]{\tiny \srclen}}} & \cellcolor{green!19} 0.12 & \cellcolor{green!53} 0.34 & \cellcolor{green!9} 0.06 & \cellcolor{green!31} 0.20 & \cellcolor{green!21} 0.13 & \cellcolor{green!53} 0.34 & \cellcolor{green!15} 0.09 & \cellcolor{green!37} 0.24 & \cellcolor{green!19} 0.12 & \cellcolor{green!56} 0.36 & \cellcolor{green!5} 0.03 & \cellcolor{green!41} 0.26 \\

\addlinespace[0.6mm]
\cdashlinelr{1-14}

\multirow[c]{3}{*}{\rotatebox[origin=c]{90}{\;\;\;\; \textbf{Internal}}} & {\colorbox{colorhex1_v2}{\makebox[1.3cm][l]{\tiny \seqprob}}} & \cellcolor{green!92} 0.59 & \cellcolor{green!30} 0.19 & \cellcolor{green!63} 0.40 & \cellcolor{green!8} 0.05 & \cellcolor{green!76} 0.48 & \cellcolor{green!26} 0.17 & \cellcolor{green!55} 0.35 & \cellcolor{green!13} 0.08 & \cellcolor{green!76} 0.48 & \cellcolor{green!24} 0.16 & \cellcolor{green!46} 0.29 & \cellcolor{green!7} 0.04 \\

 & {\colorbox{colorhex1_v2}{\makebox[1.3cm][l]{\tiny \surprisal}}} & \cellcolor{green!97} 0.62 & \cellcolor{green!32} 0.21 & \cellcolor{green!66} 0.42 & \cellcolor{green!8} 0.05 & \cellcolor{green!84} 0.53 & \cellcolor{green!31} 0.20 & \cellcolor{green!59} 0.38 & \cellcolor{green!14} 0.09 & \cellcolor{green!84} 0.53 & \cellcolor{green!28} 0.18 & \cellcolor{green!51} 0.33 & \cellcolor{green!10} 0.07 \\

 & {\colorbox{colorhex1_v2}{\makebox[1.3cm][l]{\tiny \greedyent}}} & \cellcolor{green!100} 0.64 & \cellcolor{green!32} 0.20 & \cellcolor{green!71} 0.45 & \cellcolor{green!10} 0.06 & \cellcolor{green!91} 0.58 & \cellcolor{green!33} 0.21 & \cellcolor{green!65} 0.41 & \cellcolor{green!17} 0.11 & \cellcolor{green!88} 0.56 & \cellcolor{green!31} 0.20 & \cellcolor{green!54} 0.34 & \cellcolor{green!12} 0.08 \\

\midrule
\multicolumn{14}{c}{\textbf{\towertwob}} \\
\midrule

\multirow[c]{5}{*}{\rotatebox[origin=c]{90}{\textbf{Source-only\;\;\;}}} & {\colorbox{colorhex2_v2}{\makebox[1.3cm][l]{\tiny \precomet\textsubscript{diff.}}}} & \cellcolor{green!47} 0.33 & \cellcolor{green!37} 0.26 & \cellcolor{green!40} 0.28 & \cellcolor{green!11} 0.08 & \cellcolor{green!52} 0.37 & \cellcolor{green!43} 0.30 & \cellcolor{green!38} 0.27 & \cellcolor{green!20} 0.14 & \cellcolor{green!47} 0.33 & \cellcolor{green!38} 0.27 & \cellcolor{green!33} 0.23 & \cellcolor{green!17} 0.12 \\

 & {\colorbox{colorhex2_v2}{\makebox[1.3cm][l]{\tiny \precomet\textsubscript{div.}}}} & \cellcolor{green!26} 0.18 & \cellcolor{green!35} 0.25 & \cellcolor{green!27} 0.19 & \cellcolor{green!9} 0.06 & \cellcolor{green!36} 0.25 & \cellcolor{green!43} 0.30 & \cellcolor{green!31} 0.22 & \cellcolor{green!20} 0.14 & \cellcolor{green!23} 0.16 & \cellcolor{green!30} 0.21 & \cellcolor{green!21} 0.15 & \cellcolor{green!13} 0.09 \\

 & {\colorbox{colorhex2_v2}{\makebox[1.3cm][l]{\tiny \sentinel\textsubscript{24}}}} & \cellcolor{green!35} 0.24 & \cellcolor{green!46} 0.33 & \cellcolor{green!35} 0.25 & \cellcolor{green!17} 0.12 & \cellcolor{green!37} 0.26 & \cellcolor{green!43} 0.30 & \cellcolor{green!30} 0.21 & \cellcolor{green!22} 0.15 & \cellcolor{green!35} 0.25 & \cellcolor{green!40} 0.28 & \cellcolor{green!23} 0.16 & \cellcolor{green!17} 0.12 \\

 & {\colorbox{colorhex2_v2}{\makebox[1.3cm][l]{\tiny \sentinel\textsubscript{25}}}} & \cellcolor{green!40} 0.28 & \cellcolor{green!44} 0.31 & \cellcolor{green!33} 0.23 & \cellcolor{green!14} 0.10 & \cellcolor{green!39} 0.27 & \cellcolor{green!47} 0.33 & \cellcolor{green!27} 0.19 & \cellcolor{green!19} 0.13 & \cellcolor{green!39} 0.27 & \cellcolor{green!37} 0.26 & \cellcolor{green!22} 0.15 & \cellcolor{green!13} 0.09 \\

 & {\colorbox{colorhex3_v2!80}{\makebox[1.3cm][l]{\tiny \srclen}}} & \cellcolor{red!23} -0.16 & \cellcolor{green!37} 0.26 & \cellcolor{green!3} 0.02 & \cellcolor{green!24} 0.16 & \cellcolor{red!15} -0.11 & \cellcolor{green!41} 0.29 & \cellcolor{red!3} -0.02 & \cellcolor{green!19} 0.14 & \cellcolor{red!23} -0.16 & \cellcolor{green!35} 0.25 & \cellcolor{red!17} -0.12 & \cellcolor{green!15} 0.10 \\

\addlinespace[0.6mm]
\cdashlinelr{1-14}

\multirow[c]{4}{*}{\rotatebox[origin=c]{90}{\;\;\;\; \textbf{Internal}}} & {\colorbox{colorhex1_v2}{\makebox[1.3cm][l]{\tiny \seqprob}}} & \cellcolor{green!92} 0.64 & \cellcolor{green!27} 0.19 & \cellcolor{green!47} 0.33 & \cellcolor{red!3} -0.02 & \cellcolor{green!88} 0.62 & \cellcolor{green!26} 0.18 & \cellcolor{green!57} 0.40 & \cellcolor{green!4} 0.03 & \cellcolor{green!87} 0.61 & \cellcolor{green!10} 0.07 & \cellcolor{green!63} 0.44 & \cellcolor{green!2} 0.02 \\

 & {\colorbox{colorhex1_v2}{\makebox[1.3cm][l]{\tiny \surprisal}}} & \cellcolor{green!96} 0.68 & \cellcolor{green!29} 0.21 & \cellcolor{green!50} 0.35 & \cellcolor{red!4} -0.03 & \cellcolor{green!95} 0.67 & \cellcolor{green!27} 0.19 & \cellcolor{green!63} 0.44 & \cellcolor{green!3} 0.02 & \cellcolor{green!93} 0.65 & \cellcolor{green!11} 0.08 & \cellcolor{green!65} 0.46 & \cellcolor{green!3} 0.02 \\

 & {\colorbox{colorhex1_v2}{\makebox[1.3cm][l]{\tiny \greedyent}}} & \cellcolor{green!100} 0.70 & \cellcolor{green!23} 0.16 & \cellcolor{green!51} 0.36 & \cellcolor{red!4} -0.03 & \cellcolor{green!100} 0.70 & \cellcolor{green!24} 0.17 & \cellcolor{green!65} 0.46 & \cellcolor{green!1} 0.00 & \cellcolor{green!96} 0.67 & \cellcolor{green!9} 0.07 & \cellcolor{green!68} 0.48 & \cellcolor{green!2} 0.02 \\

\bottomrule
\end{tabular}

\caption{Predictor correlations with total scaling gain
$\rho_G$ and the sampling saturation point $\rho_{N^{.95}}$ across models and
language pairs.}

\label{tab:predictor-correlations-lang-pairs}
\end{table*}

\section{Effect of decoder--evaluator alignment} \label{app:decoder-evaluator-alignment} 

The results in Section~\ref{sec:pre-search} show stronger correlations with total scaling gain for MBR than for reranking. One possible explanation is decoder--evaluator alignment: in our main experiments, MBR uses \comet{} as its utility function, while total scaling gain is also measured with \comet{}. We conduct an additional controlled experiment to isolate this effect. We use \qwenthreeeone{} and consider both \comet{} and \textsc{xComet} as MBR utility functions. For computational efficiency, we use the \textsc{xComet-Lite} variant of \textsc{xComet} \cite{larionov-etal-2024-xcomet}. Independently, we use each metric to evaluate the resulting translations and compute total scaling gain. This produces four decoder--evaluator combinations: two aligned settings, where the same metric is used for MBR and evaluation, and two mismatched settings, where decoding and evaluation use different metrics. To reduce the computational cost of running MBR with both metrics, we set the maximum candidate budget to $N_{\max}=128$ for this analysis. We otherwise follow the same procedure used to estimate total scaling gain in the main experiments.

Table~\ref{tab:metric-alignment-correlations} shows two complementary effects. First, model-internal uncertainty signals are consistently more strongly correlated with total scaling gain when the MBR utility and evaluation metric are aligned. For example, \greedyent{} reaches $\rho_G=0.49$ when both decoding and evaluation use \comet{}, compared with $0.27$ when the same \comet{}-based decoder is evaluated with \textsc{xComet}. Similarly, when MBR uses \textsc{xComet}, its correlation increases from $0.30$ under \comet{} evaluation to $0.39$ under \textsc{xComet} evaluation. The same pattern holds for \surprisal{} and \seqprob{}, indicating that the strength of the model-internal predictors in our main MBR experiments is partly amplified by decoder--evaluator alignment.

Second, under \textsc{xComet}-based evaluation, source-side signals become stronger: for example, \srclen{} reaches $\rho_G=0.33$ with \comet{} MBR and $0.49$ with \textsc{xComet} MBR, while several learned source-side predictors also become competitive with or stronger than the model-internal signals. Thus, the signals most associated with total scaling gain depend on the metric used to define search utility. Nevertheless, model-internal uncertainty remains positively correlated with total scaling gain in the mismatched settings, suggesting that it captures a signal that is not entirely specific to the decoding metric, although both its magnitude and its relative advantage over other predictors are metric-dependent.

\begin{table}[t]
\centering
\footnotesize
\setlength{\tabcolsep}{4pt}

\begin{tabular}{lcccc}
\toprule
& \multicolumn{2}{c}{\textbf{MBR {\scriptsize \comet}}}
& \multicolumn{2}{c}{\textbf{MBR {\scriptsize \textsc{xComet}}}} \\
\cmidrule(lr){2-3} \cmidrule(lr){4-5}
& \textbf{$\rho_G^{\tiny \comet}$}
& \textbf{$\rho_G^{\tiny \textsc{xComet}}$}
& \textbf{$\rho_G^{\tiny \comet}$}
& \textbf{$\rho_G^{\tiny \textsc{xComet}}$} \\
\midrule
\sentinel\textsubscript{24}       & 0.20 & 0.22 & 0.06 & 0.41 \\
\sentinel\textsubscript{25}       & 0.20 & 0.20 & 0.06 & 0.39 \\
\precomet\textsubscript{diff.}    & 0.24 & 0.20 & 0.10 & 0.35 \\
\precomet\textsubscript{div.}     & 0.31 & 0.27 & 0.19 & 0.41 \\
\srclen                            & 0.15 & 0.33 & 0.07 & 0.49 \\
\cdashlinelr{1-5}
\seqprob                           & 0.43 & 0.23 & 0.26 & 0.33 \\
\surprisal                         & 0.47 & 0.26 & 0.28 & 0.37 \\
\greedyent                         & 0.49 & 0.27 & 0.30 & 0.39 \\
\bottomrule
\end{tabular}

\caption{
Spearman correlation between each predictor and total scaling gain
$G_i^d$ for \qwenthreeeone{} averaged across language pairs with $N_{\max}=128$.
Columns vary the metric used as the MBR utility and the held-out metric
used to compute total scaling gain. The first and fourth columns correspond
to decoder--evaluator aligned settings, while the remaining columns use
different metrics for decoding and evaluation.
}
\label{tab:metric-alignment-correlations}
\end{table}

\section{Estimating marginal-gain correlations}
\label{app:marginal-estimation}

Marginal scaling gain is estimated using $R=20$ nested resampling trajectories, as described in Section~\ref{sec:formalization}. For each sentence and budget $n$, we repeatedly sample nested candidate subsets of sizes $n/2$, $n$, and $2n$ from the full candidate pool. Pool-dependent predictors are computed from the subset at budget $n$. For \objstab{}, we instead compute the absolute difference between the decoder-objective scores at budgets $n/2$ and $n$. We then average both the predictor values and the realized marginal gains across the $R$ resampling trajectories. Note that \objstab{} is first available at $n=8$, using the objectives at budgets $4$ and $8$. Spearman correlations are computed across sentences between these averaged predictor values and the corresponding estimates of marginal scaling gain. Algorithm~\ref{alg:marginal-corr} summarizes the procedure.

\begin{algorithm}[h]
\small
\caption{Estimation of marginal-gain correlations at budget $n$.}
\label{alg:marginal-corr}
\begin{algorithmic}[1]
\Require Candidate pools $\{C_i^{N_{\max}}\}_i$, decoder $d$,
evaluation metric $M$, predictor $f$, budget $n$,
number of resampling trials $R$
\For{each sentence $i$}
\For{$r=1,\ldots,R$}
\State Sample a random permutation of $C_i^{N_{\max}}$
\State Construct nested subsets of sizes $n/2$, $n$, and $2n$
\State Decode and evaluate the subsets at $n$ and $2n$
\State $\delta_i^r \gets
M(d(x_i,C_i^{2n}),z_i)
-
M(d(x_i,C_i^{n}),z_i)$
\If{$f = \mathrm{ObjStab}$}
\State Get decoder objectives
$u_i^{d}(n/2)$ and $u_i^{d}(n)$
\State $s_i^r \gets
|u_i^{d}(n)-u_i^{d}(n/2)|$
\Else
\State $s_i^r \gets f(C_i^{n})$
\EndIf
\EndFor
\State $\Delta_i^d(n,n)
\gets \frac{1}{R}\sum_{r=1}^{R}\delta_i^r$
\State $\bar{s}_i(n)
\gets \frac{1}{R}\sum_{r=1}^{R}s_i^r$
\EndFor
\State $\rho \gets
\operatorname{Spearman}
\left(
\{\bar{s}_i(n)\}_i,
\{\Delta_i^d(n,n)\}_i
\right)$
\end{algorithmic}
\end{algorithm}

\section{Additional selective search expansion results} 
\label{app:selective-expansion} 

\begin{figure*}[h]
    \centering
    \includegraphics[width=\linewidth]{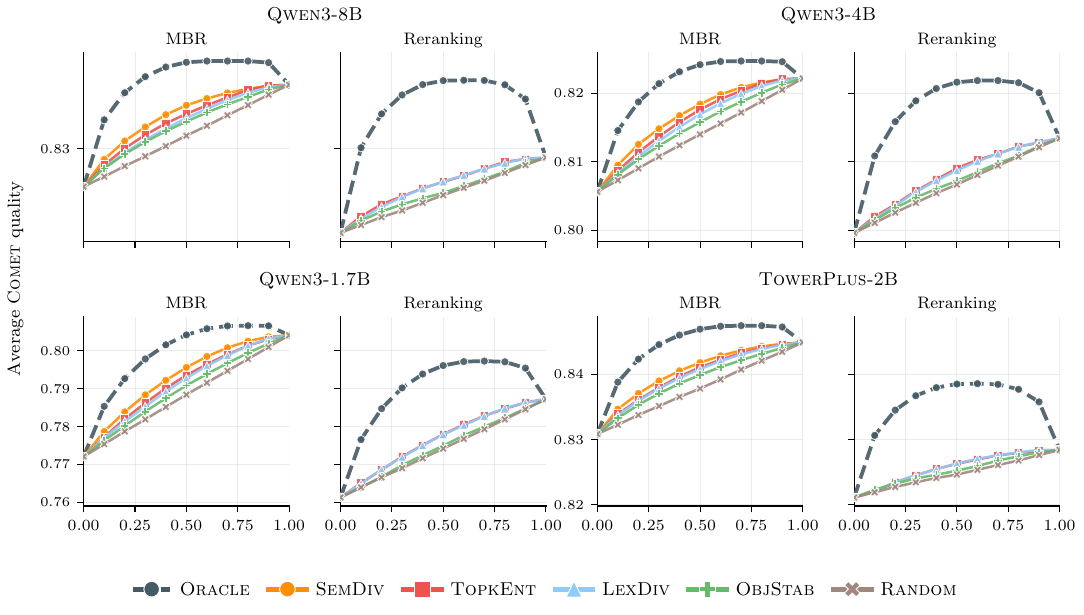}
    \caption{Final dataset-level translation quality (vertical axis) as a function of the fraction $p$ of sentences selected (horizontal axis) from $N_{\mathrm{low}}=8$ to $N_{\mathrm{high}}=128$ candidates. Sentences not selected remain at the initial budget.
    }
    \label{fig:selective-expansion1}
\end{figure*}

\begin{figure*}[h]
    \centering
    \includegraphics[width=\linewidth]{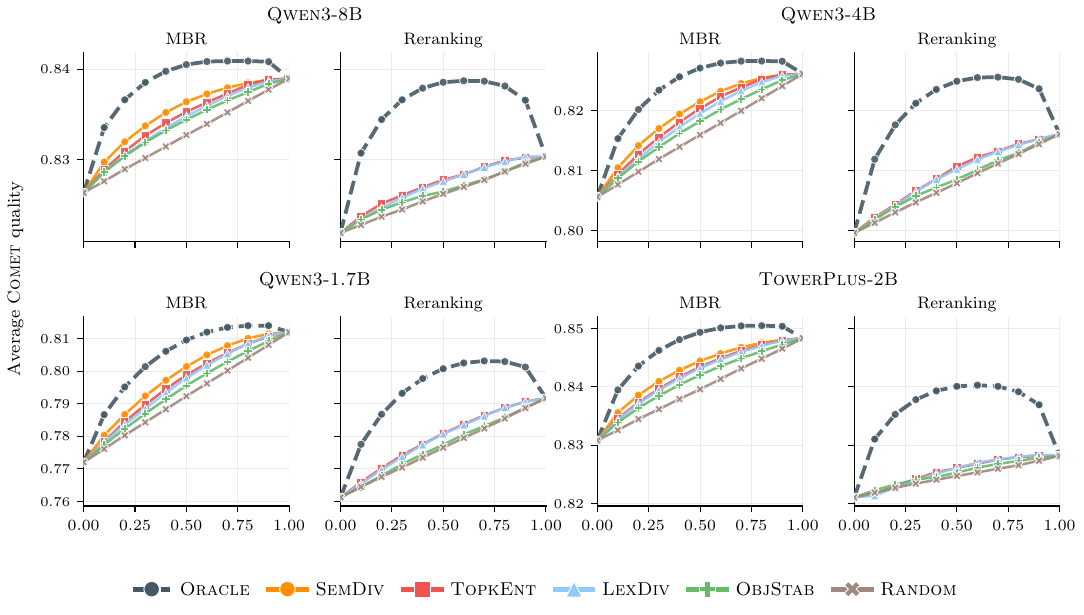}
    \caption{Final dataset-level translation quality (vertical axis) as a function of the fraction $p$ of sentences selected (horizontal axis) from $N_{\mathrm{low}}=8$ to $N_{\mathrm{high}}=512$ candidates. Sentences not selected remain at the initial budget.
    }
    \label{fig:selective-expansion2}
\end{figure*}

In Section~\ref{sec:allocation}, we evaluate selective search expansion for $N_{\mathrm{low}}=8$ and report results for $N_{\mathrm{high}}=256$ in Figure~\ref{fig:selective-expansion}. Here, we report the corresponding results for additional expansion budgets. In each case, all sentences first receive $N_{\mathrm{low}}=8$ candidates, and only the top-$p$ fraction ranked by each signal is expanded to $N_{\mathrm{high}}$. Sentences not selected for expansion remain at the initial budget. Figures~\ref{fig:selective-expansion1} and \ref{fig:selective-expansion2} report final dataset-level translation quality as a function of the fraction of sentences expanded for $N_{\mathrm{high}} = 128, 512$.

\section{Selected-set gains by model and language pair}
\label{app:selected-gain} 

Section~\ref{sec:allocation} evaluates whether different ranking signals preferentially select sentences that benefit from additional search. Here, we report the absolute \comet{} gains obtained by the selected subsets, separately for each model, language pair, and decoding method. For each sentence, the scaling gain obtained by expanding its candidate pool from $N_{\mathrm{low}}$ to $N_{\mathrm{high}}$ is

\begin{equation}
g_i^d
=
M\bigl(d(x_i,C_i^{N_{\mathrm{high}}}),z_i\bigr)
-
M\bigl(d(x_i,C_i^{N_{\mathrm{low}}}),z_i\bigr).
\end{equation}

Its expectation corresponds to the marginal scaling gain defined in Section~\ref{sec:formalization},

\begin{equation}
\mathbb{E}[g_i^d]
=
\Delta_i^d
\left(
N_{\mathrm{low}},
N_{\mathrm{high}}-N_{\mathrm{low}}
\right).
\end{equation}

Let $S_{a,p}$ denote the top-$p$ fraction of sentences selected by ranking method $a$. The average selected-set gain is then

\begin{equation}
\operatorname{Gain}_{a}^{d}(p)
=
\mathbb{E}
\left[
\frac{1}{|S_{a,p}|}
\sum_{i \in S_{a,p}} g_i^d
\right].
\end{equation}

These expectations are estimated using the same $R=20$ nested resampling trajectories used throughout the selective-expansion experiments. Larger values indicate that the ranking method assigns expansion to sentences that obtain greater quality improvements from the additional candidates. At $p=1$, all methods select the full dataset and therefore obtain the same expected gain. \textsc{Oracle} ranks sentences by their true future gain and provides an upper bound on selection under this setting, while \textsc{Random} selects the same fraction of sentences uniformly at random. Tables~\ref{tab:selected-gain-qwen8b}, \ref{tab:selected-gain-qwen4b}, \ref{tab:selected-gain-qwen1b}, and \ref{tab:selected-gain-tower2b} report results for $N_{\mathrm{low}}=8$ and $N_{\mathrm{high}}=256$. We report the mean and standard deviation across the 20 resampling trials used throughout the selective-expansion experiments.

% Requires booktabs, xcolor, and the document's \cdashlinelr command.
\begin{table*}[t]
\centering
\caption{Average selected-set gain in \comet{} for \qwenthreeethree{}. Results are reported separately for each language pair and decoding method for expansions from $N_{\mathrm{low}}=8$ to $N_{\mathrm{high}}=256$. Values show the mean $\pm$ sample standard deviation over 20 resampling trials. $p$ denotes the fraction of sentences selected for expansion.}\label{tab:selected-gain-qwen8b}
\scriptsize
\begin{tabular}{lrrrrrrrrr}
\toprule
 & \multicolumn{3}{c}{en-es} & \multicolumn{3}{c}{en-de} & \multicolumn{3}{c}{en-ru} \\
\cmidrule(lr){2-4} \cmidrule(lr){5-7} \cmidrule(lr){8-10}
 & $p=0.1$ & $p=0.5$ & $p=1$ & $p=0.1$ & $p=0.5$ & $p=1$ & $p=0.1$ & $p=0.5$ & $p=1$ \\
\midrule
\multicolumn{10}{c}{\textbf{Reranking}} \\
\midrule
{\definecolor{selectedgaincell}{HTML}{D9D9D9}\colorbox{selectedgaincell!50!white}{\makebox[1cm][l]{\scriptsize $\textsc{Oracle}$}}} & $0.08$ {\tiny $\pm 0.00$} & $0.03$ {\tiny $\pm 0.00$} & $0.01$ {\tiny $\pm 0.00$} & $0.08$ {\tiny $\pm 0.00$} & $0.03$ {\tiny $\pm 0.00$} & $0.01$ {\tiny $\pm 0.00$} & $0.10$ {\tiny $\pm 0.00$} & $0.04$ {\tiny $\pm 0.00$} & $0.01$ {\tiny $\pm 0.00$} \\
\cdashlinelr{1-10}
{\definecolor{selectedgaincell}{HTML}{EF5350}\colorbox{selectedgaincell!50!white}{\makebox[1cm][l]{\scriptsize $\textsc{TopkEnt}$}}} & $0.01$ {\tiny $\pm 0.00$} & $0.01$ {\tiny $\pm 0.00$} & $0.01$ {\tiny $\pm 0.00$} & $0.01$ {\tiny $\pm 0.01$} & $0.01$ {\tiny $\pm 0.00$} & $0.01$ {\tiny $\pm 0.00$} & $0.02$ {\tiny $\pm 0.00$} & $0.01$ {\tiny $\pm 0.00$} & $0.01$ {\tiny $\pm 0.00$} \\
{\definecolor{selectedgaincell}{HTML}{90CAF9}\colorbox{selectedgaincell!50!white}{\makebox[1cm][l]{\scriptsize $\textsc{LexDiv}$}}} & $0.01$ {\tiny $\pm 0.00$} & $0.01$ {\tiny $\pm 0.00$} & $0.01$ {\tiny $\pm 0.00$} & $0.01$ {\tiny $\pm 0.01$} & $0.01$ {\tiny $\pm 0.00$} & $0.01$ {\tiny $\pm 0.00$} & $0.02$ {\tiny $\pm 0.00$} & $0.01$ {\tiny $\pm 0.00$} & $0.01$ {\tiny $\pm 0.00$} \\
{\definecolor{selectedgaincell}{HTML}{66BB6A}\colorbox{selectedgaincell!50!white}{\makebox[1cm][l]{\scriptsize $\textsc{ObjStab}$}}} & $0.01$ {\tiny $\pm 0.00$} & $0.01$ {\tiny $\pm 0.00$} & $0.01$ {\tiny $\pm 0.00$} & $0.01$ {\tiny $\pm 0.01$} & $0.01$ {\tiny $\pm 0.00$} & $0.01$ {\tiny $\pm 0.00$} & $0.02$ {\tiny $\pm 0.01$} & $0.01$ {\tiny $\pm 0.00$} & $0.01$ {\tiny $\pm 0.00$} \\
{\definecolor{selectedgaincell}{HTML}{A1887F}\colorbox{selectedgaincell!50!white}{\makebox[1cm][l]{\scriptsize $\textsc{Random}$}}} & $0.01$ {\tiny $\pm 0.00$} & $0.01$ {\tiny $\pm 0.00$} & $0.01$ {\tiny $\pm 0.00$} & $0.01$ {\tiny $\pm 0.00$} & $0.01$ {\tiny $\pm 0.00$} & $0.01$ {\tiny $\pm 0.00$} & $0.01$ {\tiny $\pm 0.00$} & $0.01$ {\tiny $\pm 0.00$} & $0.01$ {\tiny $\pm 0.00$} \\
\midrule
\multicolumn{10}{c}{\textbf{MBR}} \\
\midrule
{\definecolor{selectedgaincell}{HTML}{D9D9D9}\colorbox{selectedgaincell!50!white}{\makebox[1cm][l]{\scriptsize $\textsc{Oracle}$}}} & $0.05$ {\tiny $\pm 0.00$} & $0.02$ {\tiny $\pm 0.00$} & $0.01$ {\tiny $\pm 0.00$} & $0.07$ {\tiny $\pm 0.00$} & $0.03$ {\tiny $\pm 0.00$} & $0.01$ {\tiny $\pm 0.00$} & $0.08$ {\tiny $\pm 0.01$} & $0.03$ {\tiny $\pm 0.00$} & $0.01$ {\tiny $\pm 0.00$} \\
\cdashlinelr{1-10}
{\definecolor{selectedgaincell}{HTML}{FF8F00}\colorbox{selectedgaincell!50!white}{\makebox[1cm][l]{\scriptsize $\textsc{SemDiv}$}}} & $0.02$ {\tiny $\pm 0.00$} & $0.01$ {\tiny $\pm 0.00$} & $0.01$ {\tiny $\pm 0.00$} & $0.03$ {\tiny $\pm 0.00$} & $0.02$ {\tiny $\pm 0.00$} & $0.01$ {\tiny $\pm 0.00$} & $0.04$ {\tiny $\pm 0.01$} & $0.02$ {\tiny $\pm 0.00$} & $0.01$ {\tiny $\pm 0.00$} \\
{\definecolor{selectedgaincell}{HTML}{EF5350}\colorbox{selectedgaincell!50!white}{\makebox[1cm][l]{\scriptsize $\textsc{TopkEnt}$}}} & $0.02$ {\tiny $\pm 0.00$} & $0.01$ {\tiny $\pm 0.00$} & $0.01$ {\tiny $\pm 0.00$} & $0.02$ {\tiny $\pm 0.00$} & $0.02$ {\tiny $\pm 0.00$} & $0.01$ {\tiny $\pm 0.00$} & $0.03$ {\tiny $\pm 0.01$} & $0.02$ {\tiny $\pm 0.00$} & $0.01$ {\tiny $\pm 0.00$} \\
{\definecolor{selectedgaincell}{HTML}{90CAF9}\colorbox{selectedgaincell!50!white}{\makebox[1cm][l]{\scriptsize $\textsc{LexDiv}$}}} & $0.01$ {\tiny $\pm 0.00$} & $0.01$ {\tiny $\pm 0.00$} & $0.01$ {\tiny $\pm 0.00$} & $0.02$ {\tiny $\pm 0.00$} & $0.02$ {\tiny $\pm 0.00$} & $0.01$ {\tiny $\pm 0.00$} & $0.03$ {\tiny $\pm 0.01$} & $0.02$ {\tiny $\pm 0.00$} & $0.01$ {\tiny $\pm 0.00$} \\
{\definecolor{selectedgaincell}{HTML}{66BB6A}\colorbox{selectedgaincell!50!white}{\makebox[1cm][l]{\scriptsize $\textsc{ObjStab}$}}} & $0.01$ {\tiny $\pm 0.00$} & $0.01$ {\tiny $\pm 0.00$} & $0.01$ {\tiny $\pm 0.00$} & $0.02$ {\tiny $\pm 0.00$} & $0.01$ {\tiny $\pm 0.00$} & $0.01$ {\tiny $\pm 0.00$} & $0.03$ {\tiny $\pm 0.01$} & $0.02$ {\tiny $\pm 0.00$} & $0.01$ {\tiny $\pm 0.00$} \\
{\definecolor{selectedgaincell}{HTML}{A1887F}\colorbox{selectedgaincell!50!white}{\makebox[1cm][l]{\scriptsize $\textsc{Random}$}}} & $0.01$ {\tiny $\pm 0.00$} & $0.01$ {\tiny $\pm 0.00$} & $0.01$ {\tiny $\pm 0.00$} & $0.01$ {\tiny $\pm 0.00$} & $0.01$ {\tiny $\pm 0.00$} & $0.01$ {\tiny $\pm 0.00$} & $0.01$ {\tiny $\pm 0.00$} & $0.01$ {\tiny $\pm 0.00$} & $0.01$ {\tiny $\pm 0.00$} \\
\bottomrule
\end{tabular}
\end{table*} % Requires booktabs, xcolor, and the document's \cdashlinelr command.
\begin{table*}[t]
\centering
\caption{Average selected-set gain in \comet{} for \qwenthreeetwo{}. Results are reported separately for each language pair and decoding method for expansions from $N_{\mathrm{low}}=8$ to $N_{\mathrm{high}}=256$. Values show the mean $\pm$ sample standard deviation over 20 resampling trials. $p$ denotes the fraction of sentences selected for expansion.}\label{tab:selected-gain-qwen4b}
\scriptsize
\begin{tabular}{lrrrrrrrrr}
\toprule
 & \multicolumn{3}{c}{en-es} & \multicolumn{3}{c}{en-de} & \multicolumn{3}{c}{en-ru} \\
\cmidrule(lr){2-4} \cmidrule(lr){5-7} \cmidrule(lr){8-10}
 & $p=0.1$ & $p=0.5$ & $p=1$ & $p=0.1$ & $p=0.5$ & $p=1$ & $p=0.1$ & $p=0.5$ & $p=1$ \\
\midrule
\multicolumn{10}{c}{\textbf{Reranking}} \\
\midrule
{\definecolor{selectedgaincell}{HTML}{D9D9D9}\colorbox{selectedgaincell!50!white}{\makebox[1cm][l]{\scriptsize $\textsc{Oracle}$}}} & $0.10$ {\tiny $\pm 0.00$} & $0.04$ {\tiny $\pm 0.00$} & $0.01$ {\tiny $\pm 0.00$} & $0.12$ {\tiny $\pm 0.01$} & $0.05$ {\tiny $\pm 0.00$} & $0.02$ {\tiny $\pm 0.00$} & $0.13$ {\tiny $\pm 0.01$} & $0.06$ {\tiny $\pm 0.00$} & $0.02$ {\tiny $\pm 0.00$} \\
\cdashlinelr{1-10}
{\definecolor{selectedgaincell}{HTML}{EF5350}\colorbox{selectedgaincell!50!white}{\makebox[1cm][l]{\scriptsize $\textsc{TopkEnt}$}}} & $0.02$ {\tiny $\pm 0.01$} & $0.01$ {\tiny $\pm 0.00$} & $0.01$ {\tiny $\pm 0.00$} & $0.02$ {\tiny $\pm 0.01$} & $0.02$ {\tiny $\pm 0.00$} & $0.02$ {\tiny $\pm 0.00$} & $0.03$ {\tiny $\pm 0.01$} & $0.03$ {\tiny $\pm 0.00$} & $0.02$ {\tiny $\pm 0.00$} \\
{\definecolor{selectedgaincell}{HTML}{90CAF9}\colorbox{selectedgaincell!50!white}{\makebox[1cm][l]{\scriptsize $\textsc{LexDiv}$}}} & $0.02$ {\tiny $\pm 0.01$} & $0.01$ {\tiny $\pm 0.00$} & $0.01$ {\tiny $\pm 0.00$} & $0.02$ {\tiny $\pm 0.01$} & $0.02$ {\tiny $\pm 0.00$} & $0.02$ {\tiny $\pm 0.00$} & $0.03$ {\tiny $\pm 0.01$} & $0.02$ {\tiny $\pm 0.00$} & $0.02$ {\tiny $\pm 0.00$} \\
{\definecolor{selectedgaincell}{HTML}{66BB6A}\colorbox{selectedgaincell!50!white}{\makebox[1cm][l]{\scriptsize $\textsc{ObjStab}$}}} & $0.02$ {\tiny $\pm 0.00$} & $0.01$ {\tiny $\pm 0.00$} & $0.01$ {\tiny $\pm 0.00$} & $0.02$ {\tiny $\pm 0.01$} & $0.02$ {\tiny $\pm 0.00$} & $0.02$ {\tiny $\pm 0.00$} & $0.02$ {\tiny $\pm 0.01$} & $0.02$ {\tiny $\pm 0.00$} & $0.02$ {\tiny $\pm 0.00$} \\
{\definecolor{selectedgaincell}{HTML}{A1887F}\colorbox{selectedgaincell!50!white}{\makebox[1cm][l]{\scriptsize $\textsc{Random}$}}} & $0.01$ {\tiny $\pm 0.00$} & $0.01$ {\tiny $\pm 0.00$} & $0.01$ {\tiny $\pm 0.00$} & $0.01$ {\tiny $\pm 0.01$} & $0.02$ {\tiny $\pm 0.00$} & $0.02$ {\tiny $\pm 0.00$} & $0.02$ {\tiny $\pm 0.01$} & $0.02$ {\tiny $\pm 0.00$} & $0.02$ {\tiny $\pm 0.00$} \\
\midrule
\multicolumn{10}{c}{\textbf{MBR}} \\
\midrule
{\definecolor{selectedgaincell}{HTML}{D9D9D9}\colorbox{selectedgaincell!50!white}{\makebox[1cm][l]{\scriptsize $\textsc{Oracle}$}}} & $0.08$ {\tiny $\pm 0.00$} & $0.03$ {\tiny $\pm 0.00$} & $0.01$ {\tiny $\pm 0.00$} & $0.10$ {\tiny $\pm 0.00$} & $0.04$ {\tiny $\pm 0.00$} & $0.02$ {\tiny $\pm 0.00$} & $0.10$ {\tiny $\pm 0.00$} & $0.05$ {\tiny $\pm 0.00$} & $0.02$ {\tiny $\pm 0.00$} \\
\cdashlinelr{1-10}
{\definecolor{selectedgaincell}{HTML}{FF8F00}\colorbox{selectedgaincell!50!white}{\makebox[1cm][l]{\scriptsize $\textsc{SemDiv}$}}} & $0.03$ {\tiny $\pm 0.00$} & $0.02$ {\tiny $\pm 0.00$} & $0.01$ {\tiny $\pm 0.00$} & $0.05$ {\tiny $\pm 0.00$} & $0.03$ {\tiny $\pm 0.00$} & $0.02$ {\tiny $\pm 0.00$} & $0.05$ {\tiny $\pm 0.01$} & $0.03$ {\tiny $\pm 0.00$} & $0.02$ {\tiny $\pm 0.00$} \\
{\definecolor{selectedgaincell}{HTML}{EF5350}\colorbox{selectedgaincell!50!white}{\makebox[1cm][l]{\scriptsize $\textsc{TopkEnt}$}}} & $0.03$ {\tiny $\pm 0.00$} & $0.02$ {\tiny $\pm 0.00$} & $0.01$ {\tiny $\pm 0.00$} & $0.04$ {\tiny $\pm 0.00$} & $0.03$ {\tiny $\pm 0.00$} & $0.02$ {\tiny $\pm 0.00$} & $0.04$ {\tiny $\pm 0.00$} & $0.03$ {\tiny $\pm 0.00$} & $0.02$ {\tiny $\pm 0.00$} \\
{\definecolor{selectedgaincell}{HTML}{90CAF9}\colorbox{selectedgaincell!50!white}{\makebox[1cm][l]{\scriptsize $\textsc{LexDiv}$}}} & $0.02$ {\tiny $\pm 0.00$} & $0.02$ {\tiny $\pm 0.00$} & $0.01$ {\tiny $\pm 0.00$} & $0.04$ {\tiny $\pm 0.01$} & $0.03$ {\tiny $\pm 0.00$} & $0.02$ {\tiny $\pm 0.00$} & $0.03$ {\tiny $\pm 0.01$} & $0.03$ {\tiny $\pm 0.00$} & $0.02$ {\tiny $\pm 0.00$} \\
{\definecolor{selectedgaincell}{HTML}{66BB6A}\colorbox{selectedgaincell!50!white}{\makebox[1cm][l]{\scriptsize $\textsc{ObjStab}$}}} & $0.02$ {\tiny $\pm 0.01$} & $0.02$ {\tiny $\pm 0.00$} & $0.01$ {\tiny $\pm 0.00$} & $0.03$ {\tiny $\pm 0.01$} & $0.03$ {\tiny $\pm 0.00$} & $0.02$ {\tiny $\pm 0.00$} & $0.03$ {\tiny $\pm 0.01$} & $0.03$ {\tiny $\pm 0.00$} & $0.02$ {\tiny $\pm 0.00$} \\
{\definecolor{selectedgaincell}{HTML}{A1887F}\colorbox{selectedgaincell!50!white}{\makebox[1cm][l]{\scriptsize $\textsc{Random}$}}} & $0.01$ {\tiny $\pm 0.00$} & $0.01$ {\tiny $\pm 0.00$} & $0.01$ {\tiny $\pm 0.00$} & $0.02$ {\tiny $\pm 0.00$} & $0.02$ {\tiny $\pm 0.00$} & $0.02$ {\tiny $\pm 0.00$} & $0.02$ {\tiny $\pm 0.00$} & $0.02$ {\tiny $\pm 0.00$} & $0.02$ {\tiny $\pm 0.00$} \\
\bottomrule
\end{tabular}
\end{table*} % Requires booktabs, xcolor, and the document's \cdashlinelr command.
\begin{table*}[t]
\centering
\caption{Average selected-set gain in \comet{} for \qwenthreeeone{}. Results are reported separately for each language pair and decoding method for expansions from $N_{\mathrm{low}}=8$ to $N_{\mathrm{high}}=256$. Values show the mean $\pm$ sample standard deviation over 20 resampling trials. $p$ denotes the fraction of sentences selected for expansion.}\label{tab:selected-gain-qwen1b}
\scriptsize
\begin{tabular}{lrrrrrrrrr}
\toprule
 & \multicolumn{3}{c}{en-es} & \multicolumn{3}{c}{en-de} & \multicolumn{3}{c}{en-ru} \\
\cmidrule(lr){2-4} \cmidrule(lr){5-7} \cmidrule(lr){8-10}
 & $p=0.1$ & $p=0.5$ & $p=1$ & $p=0.1$ & $p=0.5$ & $p=1$ & $p=0.1$ & $p=0.5$ & $p=1$ \\
\midrule
\multicolumn{10}{c}{\textbf{Reranking}} \\
\midrule
{\definecolor{selectedgaincell}{HTML}{D9D9D9}\colorbox{selectedgaincell!50!white}{\makebox[1cm][l]{\scriptsize $\textsc{Oracle}$}}} & $0.13$ {\tiny $\pm 0.01$} & $0.05$ {\tiny $\pm 0.00$} & $0.02$ {\tiny $\pm 0.00$} & $0.17$ {\tiny $\pm 0.01$} & $0.08$ {\tiny $\pm 0.00$} & $0.03$ {\tiny $\pm 0.00$} & $0.18$ {\tiny $\pm 0.01$} & $0.09$ {\tiny $\pm 0.00$} & $0.03$ {\tiny $\pm 0.00$} \\
\cdashlinelr{1-10}
{\definecolor{selectedgaincell}{HTML}{EF5350}\colorbox{selectedgaincell!50!white}{\makebox[1cm][l]{\scriptsize $\textsc{TopkEnt}$}}} & $0.04$ {\tiny $\pm 0.01$} & $0.03$ {\tiny $\pm 0.00$} & $0.02$ {\tiny $\pm 0.00$} & $0.04$ {\tiny $\pm 0.01$} & $0.04$ {\tiny $\pm 0.00$} & $0.03$ {\tiny $\pm 0.00$} & $0.05$ {\tiny $\pm 0.01$} & $0.04$ {\tiny $\pm 0.00$} & $0.03$ {\tiny $\pm 0.00$} \\
{\definecolor{selectedgaincell}{HTML}{90CAF9}\colorbox{selectedgaincell!50!white}{\makebox[1cm][l]{\scriptsize $\textsc{LexDiv}$}}} & $0.03$ {\tiny $\pm 0.01$} & $0.03$ {\tiny $\pm 0.00$} & $0.02$ {\tiny $\pm 0.00$} & $0.04$ {\tiny $\pm 0.01$} & $0.04$ {\tiny $\pm 0.00$} & $0.03$ {\tiny $\pm 0.00$} & $0.05$ {\tiny $\pm 0.01$} & $0.04$ {\tiny $\pm 0.00$} & $0.03$ {\tiny $\pm 0.00$} \\
{\definecolor{selectedgaincell}{HTML}{66BB6A}\colorbox{selectedgaincell!50!white}{\makebox[1cm][l]{\scriptsize $\textsc{ObjStab}$}}} & $0.03$ {\tiny $\pm 0.01$} & $0.02$ {\tiny $\pm 0.00$} & $0.02$ {\tiny $\pm 0.00$} & $0.03$ {\tiny $\pm 0.01$} & $0.04$ {\tiny $\pm 0.00$} & $0.03$ {\tiny $\pm 0.00$} & $0.03$ {\tiny $\pm 0.01$} & $0.03$ {\tiny $\pm 0.00$} & $0.03$ {\tiny $\pm 0.00$} \\
{\definecolor{selectedgaincell}{HTML}{A1887F}\colorbox{selectedgaincell!50!white}{\makebox[1cm][l]{\scriptsize $\textsc{Random}$}}} & $0.02$ {\tiny $\pm 0.00$} & $0.02$ {\tiny $\pm 0.00$} & $0.02$ {\tiny $\pm 0.00$} & $0.03$ {\tiny $\pm 0.01$} & $0.03$ {\tiny $\pm 0.00$} & $0.03$ {\tiny $\pm 0.00$} & $0.03$ {\tiny $\pm 0.01$} & $0.03$ {\tiny $\pm 0.00$} & $0.03$ {\tiny $\pm 0.00$} \\
\midrule
\multicolumn{10}{c}{\textbf{MBR}} \\
\midrule
{\definecolor{selectedgaincell}{HTML}{D9D9D9}\colorbox{selectedgaincell!50!white}{\makebox[1cm][l]{\scriptsize $\textsc{Oracle}$}}} & $0.11$ {\tiny $\pm 0.00$} & $0.05$ {\tiny $\pm 0.00$} & $0.02$ {\tiny $\pm 0.00$} & $0.15$ {\tiny $\pm 0.00$} & $0.08$ {\tiny $\pm 0.00$} & $0.04$ {\tiny $\pm 0.00$} & $0.16$ {\tiny $\pm 0.01$} & $0.08$ {\tiny $\pm 0.00$} & $0.04$ {\tiny $\pm 0.00$} \\
\cdashlinelr{1-10}
{\definecolor{selectedgaincell}{HTML}{FF8F00}\colorbox{selectedgaincell!50!white}{\makebox[1cm][l]{\scriptsize $\textsc{SemDiv}$}}} & $0.05$ {\tiny $\pm 0.01$} & $0.04$ {\tiny $\pm 0.00$} & $0.02$ {\tiny $\pm 0.00$} & $0.08$ {\tiny $\pm 0.01$} & $0.06$ {\tiny $\pm 0.00$} & $0.04$ {\tiny $\pm 0.00$} & $0.09$ {\tiny $\pm 0.00$} & $0.06$ {\tiny $\pm 0.00$} & $0.04$ {\tiny $\pm 0.00$} \\
{\definecolor{selectedgaincell}{HTML}{EF5350}\colorbox{selectedgaincell!50!white}{\makebox[1cm][l]{\scriptsize $\textsc{TopkEnt}$}}} & $0.05$ {\tiny $\pm 0.01$} & $0.03$ {\tiny $\pm 0.00$} & $0.02$ {\tiny $\pm 0.00$} & $0.06$ {\tiny $\pm 0.01$} & $0.06$ {\tiny $\pm 0.00$} & $0.04$ {\tiny $\pm 0.00$} & $0.06$ {\tiny $\pm 0.01$} & $0.06$ {\tiny $\pm 0.00$} & $0.04$ {\tiny $\pm 0.00$} \\
{\definecolor{selectedgaincell}{HTML}{90CAF9}\colorbox{selectedgaincell!50!white}{\makebox[1cm][l]{\scriptsize $\textsc{LexDiv}$}}} & $0.04$ {\tiny $\pm 0.01$} & $0.03$ {\tiny $\pm 0.00$} & $0.02$ {\tiny $\pm 0.00$} & $0.06$ {\tiny $\pm 0.00$} & $0.05$ {\tiny $\pm 0.00$} & $0.04$ {\tiny $\pm 0.00$} & $0.07$ {\tiny $\pm 0.01$} & $0.06$ {\tiny $\pm 0.00$} & $0.04$ {\tiny $\pm 0.00$} \\
{\definecolor{selectedgaincell}{HTML}{66BB6A}\colorbox{selectedgaincell!50!white}{\makebox[1cm][l]{\scriptsize $\textsc{ObjStab}$}}} & $0.04$ {\tiny $\pm 0.01$} & $0.03$ {\tiny $\pm 0.00$} & $0.02$ {\tiny $\pm 0.00$} & $0.05$ {\tiny $\pm 0.01$} & $0.05$ {\tiny $\pm 0.00$} & $0.04$ {\tiny $\pm 0.00$} & $0.05$ {\tiny $\pm 0.01$} & $0.05$ {\tiny $\pm 0.00$} & $0.04$ {\tiny $\pm 0.00$} \\
{\definecolor{selectedgaincell}{HTML}{A1887F}\colorbox{selectedgaincell!50!white}{\makebox[1cm][l]{\scriptsize $\textsc{Random}$}}} & $0.02$ {\tiny $\pm 0.00$} & $0.02$ {\tiny $\pm 0.00$} & $0.02$ {\tiny $\pm 0.00$} & $0.04$ {\tiny $\pm 0.01$} & $0.04$ {\tiny $\pm 0.00$} & $0.04$ {\tiny $\pm 0.00$} & $0.05$ {\tiny $\pm 0.01$} & $0.04$ {\tiny $\pm 0.00$} & $0.04$ {\tiny $\pm 0.00$} \\
\bottomrule
\end{tabular}
\end{table*} % Requires booktabs, xcolor, and the document's \cdashlinelr command.
\begin{table*}[t]
\centering
\caption{Average selected-set gain in \comet{} for \towertwob{}. Results are reported separately for each language pair and decoding method for expansions from $N_{\mathrm{low}}=8$ to $N_{\mathrm{high}}=256$. Values show the mean $\pm$ sample standard deviation over 20 resampling trials. $p$ denotes the fraction of sentences selected for expansion.}\label{tab:selected-gain-tower2b}
\scriptsize
\begin{tabular}{lrrrrrrrrr}
\toprule
 & \multicolumn{3}{c}{en-es} & \multicolumn{3}{c}{en-de} & \multicolumn{3}{c}{en-ru} \\
\cmidrule(lr){2-4} \cmidrule(lr){5-7} \cmidrule(lr){8-10}
 & $p=0.1$ & $p=0.5$ & $p=1$ & $p=0.1$ & $p=0.5$ & $p=1$ & $p=0.1$ & $p=0.5$ & $p=1$ \\
\midrule
\multicolumn{10}{c}{\textbf{Reranking}} \\
\midrule
{\definecolor{selectedgaincell}{HTML}{D9D9D9}\colorbox{selectedgaincell!50!white}{\makebox[1cm][l]{\scriptsize $\textsc{Oracle}$}}} & $0.09$ {\tiny $\pm 0.01$} & $0.03$ {\tiny $\pm 0.00$} & $0.00$ {\tiny $\pm 0.00$} & $0.09$ {\tiny $\pm 0.00$} & $0.03$ {\tiny $\pm 0.00$} & $0.01$ {\tiny $\pm 0.00$} & $0.11$ {\tiny $\pm 0.01$} & $0.04$ {\tiny $\pm 0.00$} & $0.01$ {\tiny $\pm 0.00$} \\
\cdashlinelr{1-10}
{\definecolor{selectedgaincell}{HTML}{EF5350}\colorbox{selectedgaincell!50!white}{\makebox[1cm][l]{\scriptsize $\textsc{TopkEnt}$}}} & $0.00$ {\tiny $\pm 0.01$} & $0.01$ {\tiny $\pm 0.00$} & $0.00$ {\tiny $\pm 0.00$} & $0.00$ {\tiny $\pm 0.01$} & $0.01$ {\tiny $\pm 0.00$} & $0.01$ {\tiny $\pm 0.00$} & $0.02$ {\tiny $\pm 0.01$} & $0.02$ {\tiny $\pm 0.00$} & $0.01$ {\tiny $\pm 0.00$} \\
{\definecolor{selectedgaincell}{HTML}{90CAF9}\colorbox{selectedgaincell!50!white}{\makebox[1cm][l]{\scriptsize $\textsc{LexDiv}$}}} & $0.00$ {\tiny $\pm 0.01$} & $0.01$ {\tiny $\pm 0.00$} & $0.00$ {\tiny $\pm 0.00$} & $0.00$ {\tiny $\pm 0.01$} & $0.01$ {\tiny $\pm 0.00$} & $0.01$ {\tiny $\pm 0.00$} & $0.02$ {\tiny $\pm 0.01$} & $0.02$ {\tiny $\pm 0.00$} & $0.01$ {\tiny $\pm 0.00$} \\
{\definecolor{selectedgaincell}{HTML}{66BB6A}\colorbox{selectedgaincell!50!white}{\makebox[1cm][l]{\scriptsize $\textsc{ObjStab}$}}} & $0.01$ {\tiny $\pm 0.01$} & $0.01$ {\tiny $\pm 0.00$} & $0.00$ {\tiny $\pm 0.00$} & $0.01$ {\tiny $\pm 0.01$} & $0.01$ {\tiny $\pm 0.00$} & $0.01$ {\tiny $\pm 0.00$} & $0.02$ {\tiny $\pm 0.01$} & $0.01$ {\tiny $\pm 0.00$} & $0.01$ {\tiny $\pm 0.00$} \\
{\definecolor{selectedgaincell}{HTML}{A1887F}\colorbox{selectedgaincell!50!white}{\makebox[1cm][l]{\scriptsize $\textsc{Random}$}}} & $0.01$ {\tiny $\pm 0.01$} & $0.00$ {\tiny $\pm 0.00$} & $0.00$ {\tiny $\pm 0.00$} & $0.01$ {\tiny $\pm 0.01$} & $0.01$ {\tiny $\pm 0.00$} & $0.01$ {\tiny $\pm 0.00$} & $0.01$ {\tiny $\pm 0.01$} & $0.01$ {\tiny $\pm 0.00$} & $0.01$ {\tiny $\pm 0.00$} \\
\midrule
\multicolumn{10}{c}{\textbf{MBR}} \\
\midrule
{\definecolor{selectedgaincell}{HTML}{D9D9D9}\colorbox{selectedgaincell!50!white}{\makebox[1cm][l]{\scriptsize $\textsc{Oracle}$}}} & $0.07$ {\tiny $\pm 0.00$} & $0.03$ {\tiny $\pm 0.00$} & $0.01$ {\tiny $\pm 0.00$} & $0.08$ {\tiny $\pm 0.00$} & $0.03$ {\tiny $\pm 0.00$} & $0.02$ {\tiny $\pm 0.00$} & $0.09$ {\tiny $\pm 0.00$} & $0.04$ {\tiny $\pm 0.00$} & $0.02$ {\tiny $\pm 0.00$} \\
\cdashlinelr{1-10}
{\definecolor{selectedgaincell}{HTML}{FF8F00}\colorbox{selectedgaincell!50!white}{\makebox[1cm][l]{\scriptsize $\textsc{SemDiv}$}}} & $0.03$ {\tiny $\pm 0.00$} & $0.02$ {\tiny $\pm 0.00$} & $0.01$ {\tiny $\pm 0.00$} & $0.05$ {\tiny $\pm 0.01$} & $0.02$ {\tiny $\pm 0.00$} & $0.02$ {\tiny $\pm 0.00$} & $0.05$ {\tiny $\pm 0.01$} & $0.03$ {\tiny $\pm 0.00$} & $0.02$ {\tiny $\pm 0.00$} \\
{\definecolor{selectedgaincell}{HTML}{EF5350}\colorbox{selectedgaincell!50!white}{\makebox[1cm][l]{\scriptsize $\textsc{TopkEnt}$}}} & $0.03$ {\tiny $\pm 0.00$} & $0.02$ {\tiny $\pm 0.00$} & $0.01$ {\tiny $\pm 0.00$} & $0.03$ {\tiny $\pm 0.01$} & $0.02$ {\tiny $\pm 0.00$} & $0.02$ {\tiny $\pm 0.00$} & $0.04$ {\tiny $\pm 0.01$} & $0.03$ {\tiny $\pm 0.00$} & $0.02$ {\tiny $\pm 0.00$} \\
{\definecolor{selectedgaincell}{HTML}{90CAF9}\colorbox{selectedgaincell!50!white}{\makebox[1cm][l]{\scriptsize $\textsc{LexDiv}$}}} & $0.03$ {\tiny $\pm 0.00$} & $0.02$ {\tiny $\pm 0.00$} & $0.01$ {\tiny $\pm 0.00$} & $0.03$ {\tiny $\pm 0.01$} & $0.02$ {\tiny $\pm 0.00$} & $0.02$ {\tiny $\pm 0.00$} & $0.04$ {\tiny $\pm 0.01$} & $0.03$ {\tiny $\pm 0.00$} & $0.02$ {\tiny $\pm 0.00$} \\
{\definecolor{selectedgaincell}{HTML}{66BB6A}\colorbox{selectedgaincell!50!white}{\makebox[1cm][l]{\scriptsize $\textsc{ObjStab}$}}} & $0.02$ {\tiny $\pm 0.00$} & $0.02$ {\tiny $\pm 0.00$} & $0.01$ {\tiny $\pm 0.00$} & $0.03$ {\tiny $\pm 0.01$} & $0.02$ {\tiny $\pm 0.00$} & $0.02$ {\tiny $\pm 0.00$} & $0.03$ {\tiny $\pm 0.00$} & $0.02$ {\tiny $\pm 0.00$} & $0.02$ {\tiny $\pm 0.00$} \\
{\definecolor{selectedgaincell}{HTML}{A1887F}\colorbox{selectedgaincell!50!white}{\makebox[1cm][l]{\scriptsize $\textsc{Random}$}}} & $0.01$ {\tiny $\pm 0.00$} & $0.01$ {\tiny $\pm 0.00$} & $0.01$ {\tiny $\pm 0.00$} & $0.02$ {\tiny $\pm 0.00$} & $0.02$ {\tiny $\pm 0.00$} & $0.02$ {\tiny $\pm 0.00$} & $0.02$ {\tiny $\pm 0.00$} & $0.02$ {\tiny $\pm 0.00$} & $0.02$ {\tiny $\pm 0.00$} \\
\bottomrule
\end{tabular}
\end{table*}

\end{document}